%% file: main.tex
\input{_constants}
\arxiv 

\pdfoutput=1
\documentclass[10pt,twocolumn,letterpaper]{article}
\input{cvpr_header}

\newcommand{\beginsupplement}{%
        \setcounter{table}{0}
        \renewcommand{\thetable}{S\arabic{table}}%
        \setcounter{figure}{0}
        \renewcommand{\thefigure}{S\arabic{figure}}%
     }

\makeatletter
\newcommand\refwithdefault[2]{%
  \@ifundefined{r@#1}{%
    #2%
  }{%
    \ref{#1}%
  }%
}
\makeatother

\begin{document}
\title{\paperTitle}
\author{\authorBlock}
\maketitle

\input{00_abstract}
\input{01_intro}
\input{02_related}

\input{03_method}

\input{10_conclusion}

{\small
\bibliographystyle{ieeenat_fullname}
\bibliography{11_references}
}

\ifarxiv 
    \clearpage 
    \appendix 
    \beginsupplement
    \noindent
    \pdfbookmark[0]{Supplementary Material}{sup_mat}
    {\Large\textbf{Supplementary Material} \vspace{1em}}
    \input{12_appendix}

\fi

\end{document}

%% file: _constants.tex
\def\paperID{10218}
\def\confName{CVPR}
\def\confYear{2024}

\def\paperTitle{DaReNeRF: Direction-aware Representation for Dynamic Scenes}

\def\authorBlock{
    Ange Lou\textsuperscript{1,2,}\thanks{This work was carried out during the internships of Ange Lou, Tianyu Luan, and Hao Ding at United Imaging Intelligence, Burlington, MA.}
    \quad
    Benjamin Planche\textsuperscript{1}
    \quad
    Zhongpai Gao\textsuperscript{1}
    \quad
    Yamin Li\textsuperscript{2}
    \\
    Tianyu Luan\textsuperscript{1,3,}\samethanks
    \quad
    Hao Ding\textsuperscript{1,4,}\samethanks
    \quad
    Terrence Chen\textsuperscript{1}
    \quad
    Jack Noble\textsuperscript{2}
    \quad
    Ziyan Wu\textsuperscript{1}\\
    {\normalsize
    \textsuperscript{1}United Imaging Intelligence
    \;
    \textsuperscript{2}Vanderbilt University
    \;
    \textsuperscript{3}Johns Hopkins University
    \;
    \textsuperscript{4}University of Buffalo
    }
    \\
    {\tt\small \{first.last\}@uii-ai.com, \{first.last\}@vanderbilt.edu, tianyulu@buffalo.edu, hding15@jhu.edu}
}

\newif\ifreview 
\newif\ifarxiv \newcommand{\arxiv}{\arxivtrue}
\newif\ifcamera 
\newif\ifrebuttal 

%% file: cvpr_header.tex
\ifreview \usepackage[review]{cvpr} \fi
\ifarxiv \usepackage[pagenumbers]{cvpr} \fi
\ifrebuttal \usepackage[rebuttal]{cvpr} \fi
\ifcamera \usepackage{cvpr} \fi

\input{_macros}  

\usepackage[table]{xcolor}

\usepackage{xr-hyper}

\makeatletter
\newcommand*{\addFileDependency}[1]{
  \typeout{(#1)}
  \@addtofilelist{#1}
  \IfFileExists{#1}{}{\typeout{No file #1.}}
}

\makeatother

\definecolor{cvprblue}{rgb}{0.21,0.49,0.74}
\usepackage[pagebackref,breaklinks,colorlinks,citecolor=cvprblue]{hyperref}
\usepackage[capitalize]{cleveref}
\crefname{section}{Sec.}{Secs.}
\crefname{table}{Table}{Tables}
\crefname{figure}{Fig.}{Figs.}

%% file: _macros.tex
\usepackage{graphicx}	
\usepackage{amsmath}	
\usepackage{amssymb}	
\usepackage{booktabs}
\usepackage{times}
\usepackage{microtype}
\usepackage{epsfig}
\usepackage[table,xcdraw,dvipsnames]{xcolor}
\usepackage{caption}
\usepackage{float}
\usepackage{placeins}
\usepackage{color, colortbl}
\usepackage{stfloats}
\usepackage{enumitem}
\usepackage{tabularx}
\usepackage{xstring}
\usepackage{multirow}
\usepackage{xspace}
\usepackage{url}
\usepackage{subcaption}
\usepackage{xcolor}
\usepackage[hang,flushmargin]{footmisc}
\usepackage{soul}
\usepackage{pgfplots}
\usepackage{xr}
\usepackage{animate}

\ifcamera \usepackage[accsupp]{axessibility} \fi

% ...

\ifarxiv  \fi

\newcommand{\R}[1]{{%
    \textbf{%
        \ifstrequal{#1}{1}{\textcolor{red}{R#1}}{%
        \ifstrequal{#1}{2}{\textcolor{blue}{R#1}}{%
        \ifstrequal{#1}{3}{\textcolor{magenta}{R#1}}{%
        \ifstrequal{#1}{4}{\textcolor{teal}{R#1}}{%
                           \textcolor{cyan}{R#1}%
        }}}}%
    }%
}}

\newcommand{\bigO}{\mathcal{O}}

\DeclareMathOperator{\sg}{sg}
\DeclareMathOperator{\sigmoid}{sigmoid}

\newcommand*\samethanks[1][\value{footnote}]{\footnotemark[#1]}

%% file: 00_abstract.tex
\begin{abstract}
Addressing the intricate challenge of modeling and re-rendering dynamic scenes, most recent approaches have sought to simplify these complexities using plane-based explicit representations, overcoming the slow training time issues associated with methods like Neural Radiance Fields (NeRF) and implicit representations. However, the straightforward decomposition of 4D dynamic scenes into multiple 2D plane-based representations proves insufficient for re-rendering high-fidelity scenes with complex motions. In response, we present a novel direction-aware representation (DaRe) approach that captures scene dynamics from six different directions. This learned representation undergoes an inverse dual-tree complex wavelet transformation (DTCWT) to recover plane-based information. DaReNeRF computes features for each space-time point by fusing vectors from these recovered planes. Combining DaReNeRF with a tiny MLP for color regression and leveraging volume rendering in training yield state-of-the-art performance in novel view synthesis for complex dynamic scenes. Notably, to address redundancy introduced by the six real and six imaginary direction-aware wavelet coefficients, we introduce a trainable masking approach, mitigating storage issues without significant performance decline. Moreover, DaReNeRF maintains a 2$\times$ reduction in training time compared to prior art while delivering superior performance.
\vspace{-1em}
\end{abstract}

%% file: 01_intro.tex
\section{Introduction}
\label{sec:intro}
The reconstruction and re-rendering of 3D scenes from a set of 2D images pose a fundamental challenge in computer vision, holding substantial implications for a range of AR/VR applications \cite{wang2022neural,moreau2022lens,wysocki2023ultra}. 
Despite recent progress in reconstructing static scenes, significant challenges remain. 
Real-world scenes are inherently dynamic, characterized by intricate motion, further adding to the task complexity.

Recent dynamic scene reconstruction methods build on NeRF's implicit representation. Some utilize a large MLP to process spatial and temporal point positions, generating color outputs \cite{li2022neural,li2021neural,wang2021ibrnet}. Others aim to disentangle scene motion and appearance \cite{gao2021dynamic,park2021nerfies,park2021hypernerf,pumarola2021d,liu2023robust}. However, both approaches face computational challenges, requiring extensive MLP evaluations for novel view rendering. The slow training process, often spanning days or weeks, and the reliance on additional supervision like depth maps \cite{li2021neural,li2023dynibar,liu2023robust} limit their widespread adoption for dynamic scene modeling. Several recent studies \cite{cao2023hexplane,fridovich2023k,shao2023tensor4d} have proposed decomposition-based methods to address the training time challenge. However, relying solely on decomposition limits NeRF's ability to capture high-fidelity texture details.

Recent studies \cite{rho2023masked, xu2023wavenerf, wang2022fourier, wu2023neural, yang2023freenerf} have explored the possibility of incorporating frequency information into NeRF. These frequency-based representations demonstrate promising performance in static-scene rendering, particularly in recovering detailed information. 
However, there is limited exploration \wrt the ability of these methods to scale from static to dynamic scenes. 
Additionally, HexPlane \cite{cao2023hexplane} has noted a significant degradation in reconstruction performance when using wavelet coefficients as a basis. This limitation is inherent to wavelets themselves, and we delve into a detailed discussion in the following paragraph.

Traditional 2D discrete wavelet transform (DWT) employs low/high-pass real wavelets to decompose a 2D image or grid into approximation and detail wavelet coefficients across different scales. These coefficients offer an efficient representation of both global and local image details. However, there are two significant drawbacks hindering the successful application of 2D DWT-based representations to dynamic scenes. The first is the \textbf{shift variance} problem \cite{bradley2003shift}, where even a small shift in the input signal significantly disrupts the wavelets' oscillation pattern. In dynamic 3D scenes, shifts are more pronounced than in static scenarios due to factors such as multi-object motion, camera motion, reflections, and variations in illumination. Simple DWT wavelet representations struggle to handle such 
variability, 
yielding poor results in dynamic regions.
Another critical issue is the \textbf{poor direction selectivity} \cite{kingsbury1999image} in DWT representations. A 2D DWT produces a checkered pattern that blends representations from $\pm45^\circ$, lacking directional selectivity, which is less effective for capturing lines and edges in images. Consequently, DWT-based representations fail to adequately model dynamic scenes, leading to results with noticeable ghosting artifacts around moving objects as shown in Figure \ref{teaser_img}.

This paper addresses these key limitations of the discrete wavelet transform (DWT) by introducing an efficient and robust frequency-based representation designed to overcome the challenges of shift variance and lack of direction selectivity in modeling dynamic scenes. Inspired by the dual-tree complex wavelet transform (DTCWT) \cite{selesnick2005dual}, we propose a direction-aware representation, aiming to learn features from six distinct orientations without introducing the checkerboard pattern observed in DWT. Leveraging the properties of complex wavelet transforms, our approach ensures shift invariance within the representation. The proposed direction-aware representation proves successful in modeling complex dynamic scenes, achieving state-of-the-art performance.

Furthermore, we observe that our proposed direction-aware representation introduces a $2^d$ redundancy (with $d=2$ for plane-based decomposition) compared to the DWT representation, resulting in lower storage efficiency. To address this storage challenge, we leverage a compression pipeline originally designed for static 3D scenes and adapt it for dynamic scenes. This migration of the compression pipeline proves effective in mitigating the storage constraints inherent in the direction-aware representation, making it as memory-efficient as recent state-of-the-art methods.

Additionally, to highlight the generalizability of our proposed method (aimed at 4D scenarios), we extend its application to modelling static 3D scenes. In this context, DaReNeRF demonstrates high-fidelity reconstruction performance and efficient storage capabilities. This versatility underscores the efficacy of our approach 
not only in dynamic scenes but also in static environments. This affirms its potential as a general representation utility across various scenarios.

In summary, our contributions are as follows:
\begin{itemize}
    \item We are the first to leverage DTCWT in NeRF optimization, introducing a direction-aware representation to address the shift-variance and direction-ambiguity shortcomings in DWT-based representations. DaReNeRF thereby outperforms prior decomposition-based methods in modeling complex dynamic scenes.
    \item We implement a trainable mask method for dynamic scene reconstruction, effectively resolving the storage limitations associated with the direction-aware representation. This adaptation ensures that it attains comparable memory efficiency with the current state-of-the-art methods.
    \item We extend our direction-aware representation to static scene reconstruction, and experiments demonstrate that our proposed method outperforms other state-of-the-art approaches, achieving a superior trade-off between performance and model size.
\end{itemize}

%% file: 02_related.tex
\section{Related Work}
\label{sec:related}
\noindent
\textbf{Neural Scene Representation.} NeRF \cite{mildenhall2021nerf} and its variants \cite{barron2021mip,barron2022mip,mildenhall2022nerf,niemeyer2022regnerf,low2023robust,wang2023f2,bian2023nope,yan2023nerf} show impressive results on novel view synthesis and many other application including 3D reconstruction \cite{martin2021nerf,zhang2022nerfusion,zhu2022nice,kobayashi2022decomposing}, semantic segmentation \cite{liu2023instance,mirzaei2023spin}, object detection \cite{hu2023nerf,xu2023nerf,xie2023pixel,xu2023mononerd},  generative model \cite{chan2022efficient,chan2021pi,xie2023high}, and 3D content creation \cite{metzer2023latent,wang2023learning,deng2023nerdi}.
Implicit neural representation exhibit remarkable imaging quality but suffer from slow rendering due to the numerous costly MLP evaluations required for each pixel. Numerous spatial decomposition methods \cite{fridovich2022plenoxels,chen2022tensorf,attal2023hyperreel,chan2022efficient} have been proposed to address the challenge of training speed in static scenes.

Further applying neural radiance fields to dynamic scenes is a crucial challenge. 
One straightforward approach involves extending a static NeRF by introducing an additional time dimension \cite{pumarola2021d} or latent code \cite{guo2023forward,liu2023robust,li2023dynibar,wang2023flow}. While these methods demonstrate strong capabilities in modeling complex real-world dynamic scenes, they face a severely under-constrained problem that necessitates additional supervision---\eg, depth, optical flow, and dense observations---to achieve satisfactory results. The substantial system size and weeks-long training times associated with these approaches hinder their real-world applicability. Another solution involves employing individual MLPs to represent the deformation field and a canonical field \cite{pumarola2021d,song2022pref,zhang2023deformtoon3d,yan2023nerf,johnson2023unbiased}. The latter field depicts a static scene, while the former learns coordinate mappings to the canonical space over time. Although this is an improvement over the first approach, it still requires considerable training time.

\noindent
\textbf{Scene Decomposition.} 
Recently, decomposition-based methods \cite{cao2023hexplane, fridovich2023k,shao2023tensor4d} have emerged for dynamic scenes. These approaches aim to alleviate the lengthy training times associated with dynamic scenes while maintaining the ability to model their complexity. They decompose a 4D scene into plane-based representations and employ a compact MLP to aggregate features for volumetric rendering of resulting images. While these methods significantly reduce training time and memory storage, they still encounter challenges in preserving detailed texture information during rendering. 

Wavelet-based representations \cite{rho2023masked,xu2023wavenerf,saragadam2023wire} have garnered significant attention for enhancing NeRF's ability to capture such fine texture details, owing to their capacity for recovering high-fidelity signals. However, there has been limited exploration of the potential of wavelet-based representations for dynamic scene modeling. Applying wavelet-based representations directly to plane-based methods can lead to a significant performance decay, as illustrated in Figure \ref{teaser_img}. Similar degradation is also reported by HexPlane \cite{cao2023hexplane}, highlighting the inherent limitations of wavelets, namely, shift variance and direction ambiguity. To overcome these limitations and build a more effective general dynamic NeRF, we propose a direction-aware representation, which 
preserves the ability to detect detailed textures without requiring additional supervision, achieving state-of-the-art performance in real-world dynamic scene reconstruction.

%% file: 03_method.tex
\section{Method}
\label{sec:method}
\begin{figure*}[tp]
    \centering
    \includegraphics[scale=0.48]{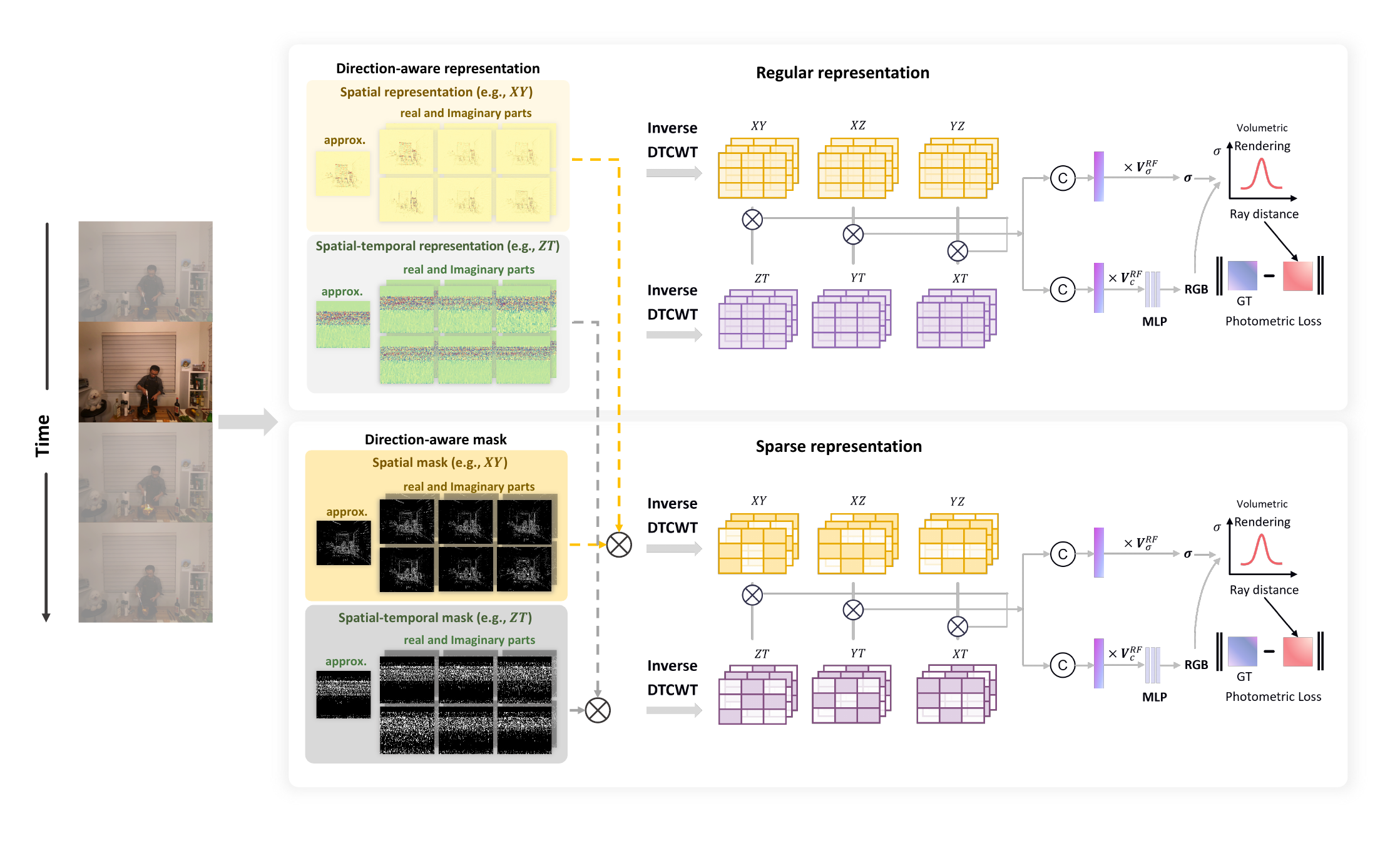}
    \caption{\textbf{Method Overview.} \textbf{Top:} The regular DaReNeRF architecture comprises an approximation and 12 direction-aware coefficient maps for both spatial (\eg, $XY$) and spatial-temporal (\eg, $ZT$) plane-based representations. To compute features of points in space-time, it multiplies feature vectors extracted from paired planes (\eg, $XY$ and $ZT$), concatenates the multiplied results into a single vector, and then multiplies them by learned tensor $V^{RF}$ for final results. RGB colors are regressed by a compact MLP, and images are synthesized via volumetric rendering. \textbf{Bottom:} The trainable mask is combined with the top architecture to create a sparse DaReNeRF. Each direction-aware representation and the approximation representation are assigned their own sparse masks. The sparse representation undergoes an inverse dual tree complex wavelet transform to obtain plane-based spatial and spatial-temporal representations.}
    \label{fig:DTCWT_method}
\vspace{-1em}
\end{figure*}
We seek to develop a model for a dynamic scene using a collection of posed images, each timestamped. The objective is to fit a model capable of rendering new images at varying poses and time stamps. Similar to D-NeRF \cite{pumarola2021d}, this model assigns color and opacity to points in both space and time. The rendering process involves differentiable volumetric rendering along rays. Training the entire model relies on a photometric loss function, comparing rendered images with ground-truth images to optimize model parameters.

Our primary innovation lies in introducing a novel direction-aware representation for dynamic scenes. This distinctive representation is coupled with the inverse dual-tree complex wavelet transform (IDTCWT) and a compact implicit multi-layer perceptron (MLP) to enable the generation of high-fidelity novel views. Figure \ref{fig:DTCWT_method} shows an overview of the model. 
Note that for simplicity, we refer to the wavelet representation as wavelet coefficients in this section.

\subsection{Dynamic Scene Decomposition}
A natural dynamic scene can be represented as a 4D spatio-temporal volume denoted as $D$. This 4D volume comprises individual static 3D volume for each time step, namely $\{V_1, V_2,...,V_T\}$. Directly modeling a 4D volume would entail a memory complexity of $\bigO(N^3TF)$, where $N$, $T$, $F$ are spatial resolution, temporal resolution and feature size (with $F=3$ representing RGB colors). To improve the overall performance, we propose a direction-aware representation applied to baseline plane-based 4D volume decomposition \cite{cao2023hexplane}. In such baseline, a representation of the 4D volume can be represented as follows:
\begin{small}
\begin{equation}
\begin{split}
    D=&\sum_{r=1}^{R_1}{M_r^{XY}\circ M_r^{ZT}\circ v_r^1}+\sum_{r=1}^{R_2}{M_r^{XZ}\circ M_r^{YT}\circ v_r^2} \\
    &+\sum_{r=1}^{R_3}{M_r^{YZ}\circ M_r^{XT}\circ v_r^3}
\end{split}
\end{equation}
\end{small}
where each $M_r^{AB}\in\mathbb{R}^{AB}$ represents a learned 2D plane-based representation with $\big\{(A,B) \in \{X, Y, Z, T\}^2 \mid A \ne B\big\}$", and $v_r^{i}\in\mathbb{R}^{F}$ are learned vectors along $F$ axes. The parameters $R_1$, $R_2$ and $R_3$ correspond to the number of low rank components. By defining  $R=R_1+R_2+R_3 \ll N$, the model's memory complexity can be notably reduced from $\bigO(N^3TF)$ to $\bigO(RN^2TF)$. This reduction in memory requirements proves advantageous for efficiently modeling dynamic scenes while preserving computational resources. 

To compute the density and appearance features of points in space-time, the model multiplies the feature vectors extracted from paired planes (\eg, $XY$ and $ZT$), concatenates the multiplied results into a single vector, and then multiplies them by $V^{RF}$, which stacks all $v_r^i$ into a 2D tensor. The point opacities are directly queried from the density features. The RGB color values are regressed by a compact MLP, where the inputs are appearance features and view directions. Finally, images are synthesized via volumetric rendering. To improve the overall performance, we apply our proposed direction-aware representation to this baseline.

\subsection{Direction-Aware Representation}
\begin{figure}
    \centering
    \includegraphics[width=1.\linewidth]{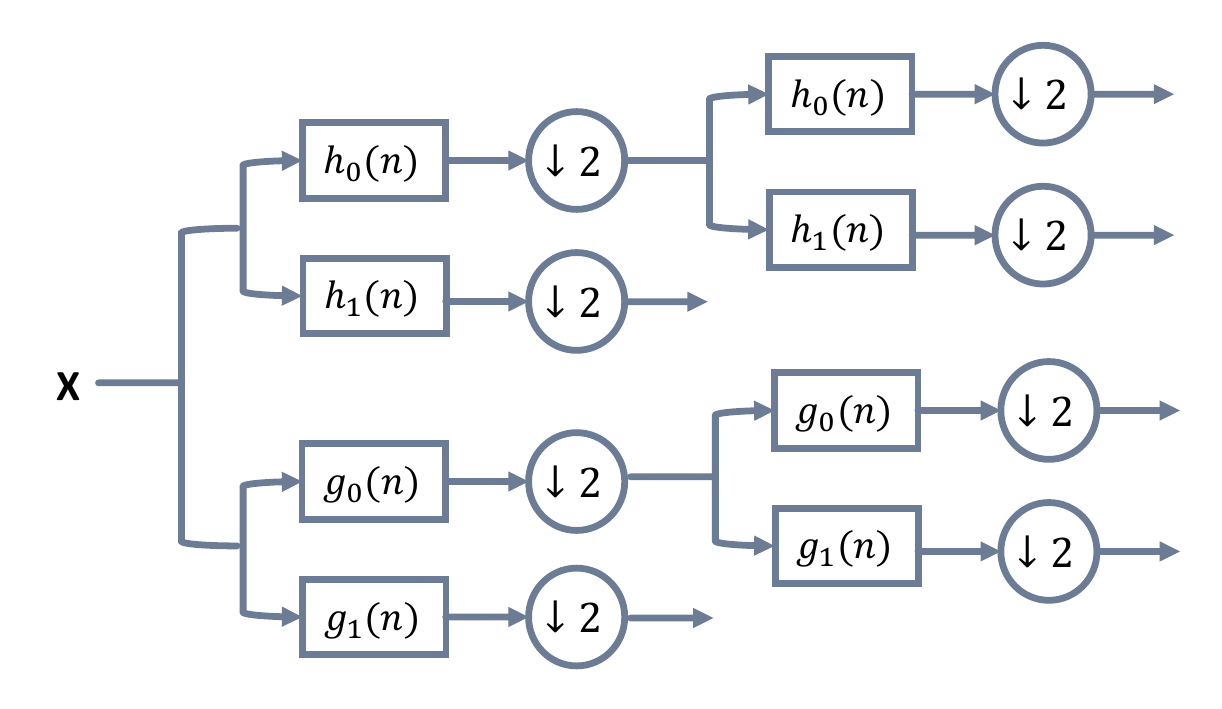}
    \caption{\textbf{Analysis Filter Bank}, for the dual tree complex wavelet transform.}
    \label{fig:dtcwt_filter_bank}
\vspace{-2em}
\end{figure}

Built upon plane-based 4D volume decomposition and drawing inspiration from the dual-tree complex wavelet transform, we introduce the direction-aware representation. This innovative approach enables the modeling of representations from six different directions. In contrast to the prevalent use of 2D discrete wavelet transforms (DWT), the dual tree complex wavelet transform (DTCWT) \cite{selesnick2005dual} employs two complex wavelets as illustrated in Figure \ref{fig:dtcwt_filter_bank}.
Given $h=[h_0,h_1]$ and $g=[g_0,g_1]$ low/high pass filter pairs for upper (real) and lower (imaginary) filter banks, the low-pass and high-pass complex wavelet transforms in DTCWT are denoted as $\phi(x)=\phi_h(x)+j\phi_g(x)$ and $\psi(x)=\psi_h(x)+j\psi_g(x)$.
Consequently, applying low- and high-pass complex wavelet transforms to rows and columns of a 2D grid yields wavelet coefficients $\phi(x)\psi(y)$, $\psi(x)\phi(y)$ and $\psi(x)\psi(y)$. Due to filter design, the upper (real) filter and lower filter (imaginary) satisfy the Hilbert transform, denoted as $\psi_g(x)\approx\mathcal{H}(\psi_h(x))$. Finally, three additional wavelet coefficients,  $\phi(x)\overline{\psi(y)}$, $\psi(x)\overline{\phi(y)}$ and $\psi(x)\overline{\psi(y)}$, can be obtained, where $\overline{\phi}$ and $\overline{\psi}$ represent the complex conjugate of $\phi$ and $\psi$.  From these 2D wavelet coefficients, we derive six direction-aware real and imaginary wavelet coefficients, each with the same set of six directions. Compared to 2D DWT, the six wavelet coefficients align along specific directions, eliminating the checkerboard effect, with more results in the supplementary material.
Exploiting the properties of DTCWT, we aim for the plane-based representation $M_r^{AB}\in\mathbb{R}^{m,n}$ of the 4D volume to possess direction-aware capabilities. Here, $m$ and $n$ denote the resolution of the 2D plane-based representation. To imbue each 2D plane-based representation with direction-aware capabilities, we introduce twelve learned wavelet coefficients—six for the real part and six for the imaginary part—denoted as $\mathbf{R}\{\mathcal{W}_i^{AB}\}_{i=1}^6 \in\mathbb{R}^{m/2^l,n/2^l}$ and $\mathbf{I}\{\mathcal{W}_i^{AB}\}_{i=1}^6 \in\mathbb{R}^{m/2^l,n/2^l}$, respectively. Additionally, a learned approximation 
coefficient is defined as $\mathcal{W}_a^{AB}\in\mathbb{R}^{m/2^{l-1},n/2^{l-1}}$, with 
$l$ the DTCWT transformation level. Consequently, a specific plane-based representation can be expressed as:
\begin{small}
\begin{equation}
    M_r^{AB} = IDTCWT([W_{a,r}^{AB},\mathbf{R}\{\mathcal{W}_{i,r}^{AB}\}_{i=1}^6,\mathbf{I}\{\mathcal{W}_{i,r}^{AB}\}_{i=1}^6])
\end{equation}
\end{small}
Importantly, our 
representation is not only applicable for modeling dynamic 3D scenes but is also well-suited for static 3D scenes, following a TensorRF-like \cite{chen2022tensorf} decomposition:
\begin{small}
\begin{equation}
\begin{split}
    D=&\sum_{r=1}^{R_1}{M_r^{XY}\circ v_r^{Z}\circ v_r^1} +\sum_{r=1}^{R_2}{M_r^{XZ}\circ v_r^{Y}\circ v_r^2} \\
    &+ \sum_{r=1}^{R_3}{M_r^{YZ}\circ v_r^{X}\circ v_r^3}
\end{split}
\end{equation} 
\end{small}
In this formulation, a plane-based representation $M_r^{AB}\in \mathbb{R}^{AB}$ and a vector-based representation $v_r^C\in\mathbb{R}^C$ are employed to model a 3D volume. For static scenes, our direction-aware representations also could be applied to represent the plane-based representations. 

\subsection{Sparse Representation and Model Compression}
In contrast to the classical 2D discrete wavelet transform (DWT), our direction-aware representation excels in modeling dynamic 3D scenes. However, it is worth noting that a single-level dual tree complex wavelet transform (DTCWT) necessitates six real direction-aware wavelet coefficients and six imaginary direction-aware wavelet coefficients to impart directional information to the plane-based representation. In contrast, a single-level 2D DWT only has three real wavelet coefficients, albeit with inherent direction ambiguity. To enhance the storage efficiency of our solution, we employ learned masks \cite{rho2023masked} for each directional wavelet coefficient, selectively masking out less important features.

To address the $2^d$ redundancies, where $d=2$ for the 2D DTCWT transform, we employ learned masks $\mathbf{R}\{\mathcal{M}_i^{AB}\}_{i=1}^6 \in\mathbb{R}^{m/2^l,n/2^l}$, $\mathbf{I}\{\mathcal{M}_i^{AB}\}_{i=1}^6 \in\mathbb{R}^{m/2^l,n/2^l}$ and $\mathcal{M}_a^{AB}\in\mathbb{R}^{m/2^{l-1},n/2^{l-1}}$ for the six real wavelet coefficients, six imaginary wavelet coefficients and the approximation coefficients, respectively. The masked wavelet coefficients can be denoted as:
\begin{equation}
    \widehat{\mathcal{W}}^{AB}=\sg\Big(\big(\mathbf{H}(\mathcal{M}^{AB})-\sigmoid(\mathcal{M}^{AB})\big)\odot\mathcal{W}^{AB}\Big)
\end{equation}
here $\big\{\mathbf{R}\{\mathcal{M}_i^{AB}\}_{i=1}^6,\mathbf{I}\{\mathcal{M}_i^{AB}\}_{i=1}^6,\mathcal{M}_a^{AB}\big\}\in\mathcal{M}^{AB}$ and $\big\{\mathbf{R}\{\mathcal{W}_i^{AB}\}_{i=1}^6,\mathbf{I}\{\mathcal{W}_i^{AB}\}_{i=1}^6,\mathcal{W}_a^{AB}\big\}\in\mathcal{W}^{AB}$. The functions $\sg$, $\mathbf{H}$ and $\sigmoid$ represent the stop-gradient operator, Heaviside step and element-wise sigmoid function, respectively. The masked plane-based representation is obtained from the masked wavelet coefficients through the equation:
\begin{small}
\begin{equation}
    \widehat{M}_r^{AB} = IDTCWT([\widehat{W}_{a,r}^{AB},\mathbf{R}\{\widehat{\mathcal{W}}_{i,r}^{AB}\}_{i=1}^6,\mathbf{I}\{\widehat{\mathcal{W}}_{i,r}^{AB}\}_{i=1}^6])
\end{equation}
\end{small}
To encourage sparsity in the generated masks, we introduce an additional loss term $\mathcal{L}_m$, defined as the sum of all masks. We employ $\lambda_m$ as the weight of $\mathcal{L}_m$ to control the sparsity of the representation.

Following the removal of unnecessary representations through masking, we adopt a compression strategy akin to the one employed in masked wavelet NeRF \cite{rho2023masked}, originally designed for static scenes, to compress the sparse representation and masks that identify non-zero elements. The process involves converting the binary mask values to 8-bit unsigned integers and subsequently applying run-length encoding (RLE). Finally, the Huffman encoding algorithm is employed on the RLE-encoded streams to efficiently map values with a high probability to shorter bits.
\subsection{Optimization}
We leverage our proposed direction-aware representation to effectively represent 3D dynamic scenes. The model is then optimized through a photometric loss function, which measures the difference between rendered images and target images. For a given point $(x,y,z,t)$, its opacity and appearance features are represented by six real and six imaginary direction-aware representation. The final color is regressed through a small multi-layer perceptron (MLP), taking the appearance feature and view direction as inputs. Utilizing the point's opacities and colors, images are obtained through volumetric rendering. 
The overall loss is expressed as:
\begin{equation}
    \mathcal{L} = \frac{1}{|\mathcal{R}|}\sum_{r\in\mathcal{R}}{||\mathbf{C}(r)-\widehat{\mathbf{C}}|(r)||_2^2} + \lambda_{reg}\mathcal{L}_{reg} + \lambda_m\mathcal{L}_m,
\end{equation}
with $\mathcal{L}_{reg}$, $\lambda_{reg}$ and $\mathcal{L}_{m}$, $\lambda_{m}$ the regularization loss and mask loss with respective weights, $\mathcal{R}$ the set of rays, and $\mathbf{C}(r)$, $\widehat{\mathbf{C}}(r)$ the rendered and ground-truth ray colors.

\noindent
\textbf{Regularization}. For the regularization term, we employ the total variational (TV) loss on the direction-aware representation to enforce spatio-temporal continuity. 

\noindent
\textbf{Training Strategy}. We employ the same coarse-to-fine training strategy as in  \cite{cao2023hexplane,chen2022tensorf,yu2021plenoctrees}, where the resolution of grids progressively increases during training. This strategy not only accelerates the training process but also imparts an implicit regularization on nearby grids.

\noindent
\textbf{Emptiness Voxel}. We maintain a small 3D voxel representation that indicates the emptiness of specific regions in the scene, allowing us to skip points located in empty regions. Given the typically large number of empty regions, this strategy significantly aids in acceleration. To generate this voxel, we evaluate the opacities of points across different time steps and aggregate them into a single voxel by retaining the maximum opacities. While preserving multiple voxels for distinct time intervals could potentially enhance efficiency, for the sake of simplicity, we opt to keep only one voxel \cite{cao2023hexplane}.

\begin{table*}[ht]
    \centering
    \setlength{\tabcolsep}{3.5mm}
    \resizebox{\linewidth}{!}{
    \begin{tabular}{rccccccc}
        \toprule
         & Model&Steps&PSNR$\uparrow$&D-SSIM$\downarrow$&LPIPS$\downarrow$&Training Time$\downarrow$&Model Size (MB) $\downarrow$\\
         \midrule
         \parbox[t]{0mm}{\multirow{8}{*}{\rotatebox[origin=c]{90}{{\footnotesize\texttt{flame-salmon}} scene}}} &
         Neural Volumes \cite{lombardi2019neural}&-&22.800&0.062&0.295&-\\
         & LLFF \cite{mildenhall2019local}&-&23.239&0.076&0.235&-&-\\
         & NeRF-T \cite{li2022neural}&-&28.449&0.023&0.100&-&-\\
         & DyNeRF \cite{li2022neural}&650k&29.581&0.020&0.099&1,344h&\textbf{28}\\
         & HexPlane \cite{cao2023hexplane}&650k&29.470&0.018&\textbf{0.078}&12h& 252\\
         & HexPlane \cite{cao2023hexplane}&100k&29.263&0.020&0.097&\textbf{2h}& 252\\
         & DaReNeRF-S&100k&\underline{30.224}&\underline{0.015}&0.089&5h& \underline{244}\\
         & DaReNeRF&100k&\textbf{30.441}&\textbf{0.012}&\underline{0.084}&\underline{4.5h}& 1,210 \\
         \midrule
         \parbox[t]{0mm}{\multirow{11}{*}{\rotatebox[origin=c]{90}{all scenes (average)}}} &
         NeRFPlayer \cite{song2023nerfplayer}&-&30.690&0.034&0.111&6h&-\\
         & HyperReel \cite{attal2023hyperreel}&-&31.100&0.036&0.096&9h&-\\
         & HexPlane \cite{cao2023hexplane}&650k&31.705&\underline{0.014}&\underline{0.075}&12h& 252\\
         & HexPlane \cite{cao2023hexplane}&100k&31.569&0.016&0.089&\underline{2h}& 252\\
         & K-Planes-explicit \cite{fridovich2023k}&120k&30.880&-&-&3.7h& 580\\
         & K-Planes-hybrid&90k&31.630&-&-&1.8h& 310\\
         & Mix Voxels-L \cite{wang2023mixed}&25k&31.340&0.019&0.096&\textbf{1.3h}& 500\\
         & Mix Voxels-X \cite{wang2023mixed}&50k&31.730&0.015&\textbf{0.064}&5h& 500\\
         & 4D-GS \cite{wu20234d}&-&31.020&-&0.150&\underline{2h}&\textbf{145}\\
         & DaReNeRF-S&100k&\underline{32.102}&\underline{0.014}&0.087&5h& \underline{244}\\
         & DaReNeRF&100k&\textbf{32.258}&\textbf{0.012}&0.084&4.5h& 1,210\\
         \bottomrule
         
    \end{tabular}
    }
    
    \caption{\textbf{Quantitative Comparison on Plenoptic Video Data.} We present results on synthesis quality and training time (measured in GPU hours).
    Following prior art, we provide both scene-specific performance ({\footnotesize\texttt{flame-salmon}} scene) and mean performance across all cases from their original papers.}
    \label{tab:plenoptic_table}
\vspace{-1em}
\end{table*}

\section{Experiments}
We evaluate the performance of our proposed direction-aware representation on both dynamic and static scenes, conducting a thorough comparison with 
prior art.
Additionally, we delve into the advantages of our direction-aware representation through ablation studies, showcasing its robustness in handling both dynamic and static scenes.
\subsection{Novel View Synthesis of Dynamic Scenes}
For dynamic scenes, we employ two distinct datasets with varying settings. 
Each dataset presents its own challenges, effectively addressed by our direction-aware representation.

\noindent
\textbf{Plenoptic Video Dataset} \cite{li2022neural} is a real-world dataset captured by a multi-view camera system using 21 GoPro cameras at a resolution of $2028 \times 2704$ and a frame rate of 30 frames per second. Each scene consists of 19 synchronized, 10-second videos, with 18 videos designated for training and one for evaluation. This dataset serves as an ideal testbed to assess the representation ability, featuring complex and challenging dynamic content, including highly specular, translucent, and transparent objects, topology changes, moving self-casting shadows, fire flames, strong view-dependent effects for moving objects, and more.

For a fair and direct comparison, we adhere to the same training and evaluation protocols as DyNeRF \cite{li2022neural}. Our model is trained on a single A100 GPU, utilizing a batch size of 4,096. We adopt identical importance sampling strategies and hierarchical training techniques as DyNeRF, employing a spatial grid size of 512 and a temporal grid size of 300. The scene is placed under the normalized device coordinates (NDC) setting, consistent with the approach outlined in \cite{mildenhall2021nerf}.

Quantitative compression results with state-of-the-art methods are presented in Table \ref{tab:plenoptic_table}. We utilize measurement metrics PSNR, structure dissimilarity index measure (DSSIM) \cite{sara2019image}, and perception quality measure LPIPS \cite{zhang2018unreasonable} to conduct a comprehensive evaluation. As demonstrated in Table \ref{tab:plenoptic_table}, leveraging the proposed direction-aware representation, both regular and sparse DaReNeRF achieve promising results compared to the most recent state-of-the-art
, with analogous training time. This more ideal trade-off between performance and computational requirements, compared to prior art, is also illustrated in Figure \ref{teaser_img}.b, computed over Plenoptic data.
Figure \ref{fig:Visual_results_plenoptic} presents some novel-view  results on the Plenoptic dataset. Four small patches, each containing detailed texture information, are selected for comparison. DaReNeRF, equipped with the proposed direction-aware representation, excels in reconstructing moving objects (\eg, dog and firing gun) and capturing better texture details (\eg, hair and metal rings on the apron).

\begin{figure*}[t]
    \centering
  \animategraphics[width=\linewidth,controls,autoplay,loop,buttonsize=.7em]{5}{img/plenoptic_anim/Slide}{1}{6}
  
    \caption{
    \textbf{Visual Comparison on Dynamic Scenes (Plenoptic Data).}
    K-Planes and HexPlane are concurrent decomposition-based methods. 
    As shown in the four zoomed-in patches, our method better reconstruct fine details and captures motion.
    To see the figure animated, please view the document with compatible software, \eg,  \textit{Adobe Acrobat} or \textit{KDE Okular.}
    }
    \label{fig:Visual_results_plenoptic}
\vspace{-1em}
\end{figure*}

\noindent
\textbf{D-NeRF Dataset} \cite{pumarola2021d} is a monocular video dataset with $360^\circ$ observations of synthetic objects. Dynamic 3D reconstruction from monocular video poses challenges as only one observation is available for each timestamp. State-of-the-art methods for monocular video typically incorporate a deformation field. 
However, the underlying assumption is that the scenes undergo no topological changes,
making them less effective for real-world cases (\eg, Plenoptic dataset). 
Table \ref{tab:dnerf_table} reports the rendering quality of different methods with and without deformation fields on the D-NeRF data, 
DaReNeRF outperforms all non-deformation methods, as well as some deformation methods, \eg D-NeRF and TiNeuVox-S \cite{fang2022fast}. 
The superiority of our solution on topologically-changing scenes is further highlighted in annex.

\begin{table}[t]
    \centering
    \caption{
    \textbf{Quantitative Study on D-NeRF Data.}
    Without 
    the topological constraints of using deformation fields, 
    DaReNeRF outperforms even some deformation-based methods.}
    \vspace{-.5em}
    \resizebox{\linewidth}{!}{
    \begin{tabular}{ccccc}
    \toprule
         Model&Deform.&PSNR$\uparrow$&SSIM$\uparrow$&LPIPS$\downarrow$  \\
         \midrule
         D-NeRF \cite{pumarola2021d}&\checkmark&30.50&0.95&0.07\\
         TiNeuVox-S \cite{fang2022fast}&\checkmark&30.75&0.96&0.07\\
         TiNeuVox-B \cite{fang2022fast}&\checkmark&\underline{32.67}&\underline{0.97}&\underline{0.04}\\
         4D-GS \cite{wu20234d}&\checkmark&\textbf{33.30}&\textbf{0.98}&\textbf{0.03}\\
         \midrule
         T-NeRF \cite{pumarola2021d}&-&29.51&\underline{0.95}&0.08\\
         HexPlane \cite{cao2023hexplane}&-&31.04&\textbf{0.97}&\underline{0.04}\\
         K-Planes \cite{fridovich2023k}&-&31.05&\textbf{0.97}&-\\
         DaReNeRF-S&-&\underline{31.82}&\textbf{0.97}&\textbf{0.03}\\
         DaReNeRF&-&\textbf{31.95}&\textbf{0.97}&\textbf{0.03}\\
         \bottomrule
    \end{tabular}
    }
    \label{tab:dnerf_table}
\vspace{-1.5em}
\end{table}

\subsection{Novel View Synthesis of Static Scenes}
For static scenes, we test our proposed direction-aware representation on NeRF synthetic \cite{mildenhall2021nerf}, Neural Sparse Voxel Fields (NSVF) \cite{liu2020neural} and LLFF \cite{mildenhall2019local} datasets. We use TensoRF-192 as baseline and apply our proposed representation. We report the performance on these three datasets in Tables \ref{tab:nerf_table}, \ref{tab:NVSF_table}, and \ref{tab:LLFF_table} respectively.

\begin{table}[t]
    \centering
    \caption{
    \textbf{Quantitative Comparison on NeRF Synth.}, with models designed for different bit-precisions ($^\ast$ denotes a model quantized post-training;  numbers in brackets denote grid resolutions).}
    \vspace{-.75em}
    \label{tab:nerf_table}
    \setlength{\tabcolsep}{3.5mm}
    \resizebox{\linewidth}{!}{
    \begin{tabular}{cccc}
        \toprule
         Precision&Method& Size (MB) & PSNR $\uparrow$\\
        \midrule
         32-bit&KiloNeRF \cite{reiser2021kilonerf}& $\leq$ 100& 31.00\\
         32-bit&CCNeRF (CP) \cite{tang2022compressible}& 4.4 & 30.55\\
         8-bit$^\ast$&NeRF \cite{mildenhall2021nerf}&1.25&31.52\\
         8-bit&cNeRF \cite{bird20213d}&\textbf{0.70}&30.49\\
         8-bit&PREF \cite{huang2022pref}&9.88&31.56\\
         8-bit$^\ast$&VM-192 \cite{chen2022tensorf} & 17.93&\textbf{32.91}\\
         8-bit$^\ast$&VM-192 (300) + DWT \cite{rho2023masked}&\underline{0.83}&31.95\\
         \midrule
         8-bit$^\ast$&VM-192 (300) + Ours &8.91&\underline{32.42}\\
         \bottomrule
         
    \end{tabular}
    }
\vspace{-.75em}
\end{table}

\begin{table}[t]
    \centering
    \caption{
    \textbf{Quantitative Comparison on NSVF} (static scenes).}
    \vspace{-.5em}
    \setlength{\tabcolsep}{3.5mm}
    \resizebox{\linewidth}{!}{
    \begin{tabular}{cccc}
    \toprule
         Bit Precision&Model & Size (MB) & PSNR $\uparrow$  \\
         \midrule
         32-bit&KiloNeRF \cite{reiser2021kilonerf}& $\leq$ 100& 33.37\\ 
         8-bit$^\ast$&VM-192 \cite{song2022pref} & 17.77 &\underline{36.11}\\
         8-bit$^\ast$&VM-48  \cite{chen2022tensorf}& 4.53 &34.95\\
         8-bit$^\ast$&CP-384 \cite{chen2022tensorf}& \textbf{0.72} &33.92\\
         8-bit$^\ast$&VM-192 (300) + DWT \cite{rho2023masked}&\underline{0.87}&34.67\\
         \midrule
         8-bit$^\ast$&VM-192 (300) + Ours &8.98&\textbf{36.24}\\
         \bottomrule
    \end{tabular}
    }
    \label{tab:NVSF_table}
\vspace{-.75em}
\end{table}

\begin{table}[t]
    \centering
    \caption{
    \textbf{Quantitative Comparison on LLFF} (static scenes).}
    \vspace{-.75em}
    \setlength{\tabcolsep}{3.5mm}
    \resizebox{\linewidth}{!}{
    \begin{tabular}{cccc}
    \toprule
         Bit Precision&Model & Size(MB) & PSNR $\uparrow$  \\
         \midrule
         8-bit&cNeRF \cite{bird20213d}& \underline{0.96}&26.15\\
         8-bit$^\ast$&PREF \cite{song2022pref} &9.34&24.50\\
         8-bit$^\ast$&VM-96  \cite{chen2022tensorf}&44.72&\textbf{26.66}\\
         8-bit$^\ast$&VM-48  \cite{chen2022tensorf}&22.40&26.46\\
         8-bit$^\ast$&CP-384 \cite{chen2022tensorf}&\textbf{0.64}&25.51\\
         8-bit$^\ast$&VM-96 (640) + DWT \cite{rho2023masked}&1.34&25.88\\
         \midrule
         8-bit$^\ast$&VM-96 (640) + Ours&13.67&\underline{26.48}\\
         \bottomrule
    \end{tabular}
    }
    
    \label{tab:LLFF_table}
\vspace{-1.5em}
\end{table}

Across these three static datasets, our direction-aware representation outperforms most compression-based NeRF models with model sizes ranging from 8 to 14MB. While our method's model size is larger than DWT-based solutions, it achieves comparable sparsity. 
For instance, with $\lambda_m=2.5\times 10^{-11}$, its \textit{sparsity} reaches 94$\%$, closely aligned with the 97$\%$ reported in the masked wavelet NeRF \cite{rho2023masked} paper. Notably, with similar sparsity, our direction-aware method exhibits PSNR improvements of 0.47, 1.57, and 0.60 over DWT-based methods on the three static datasets.

\begin{figure}
    \centering
    \includegraphics[width=1.\linewidth]{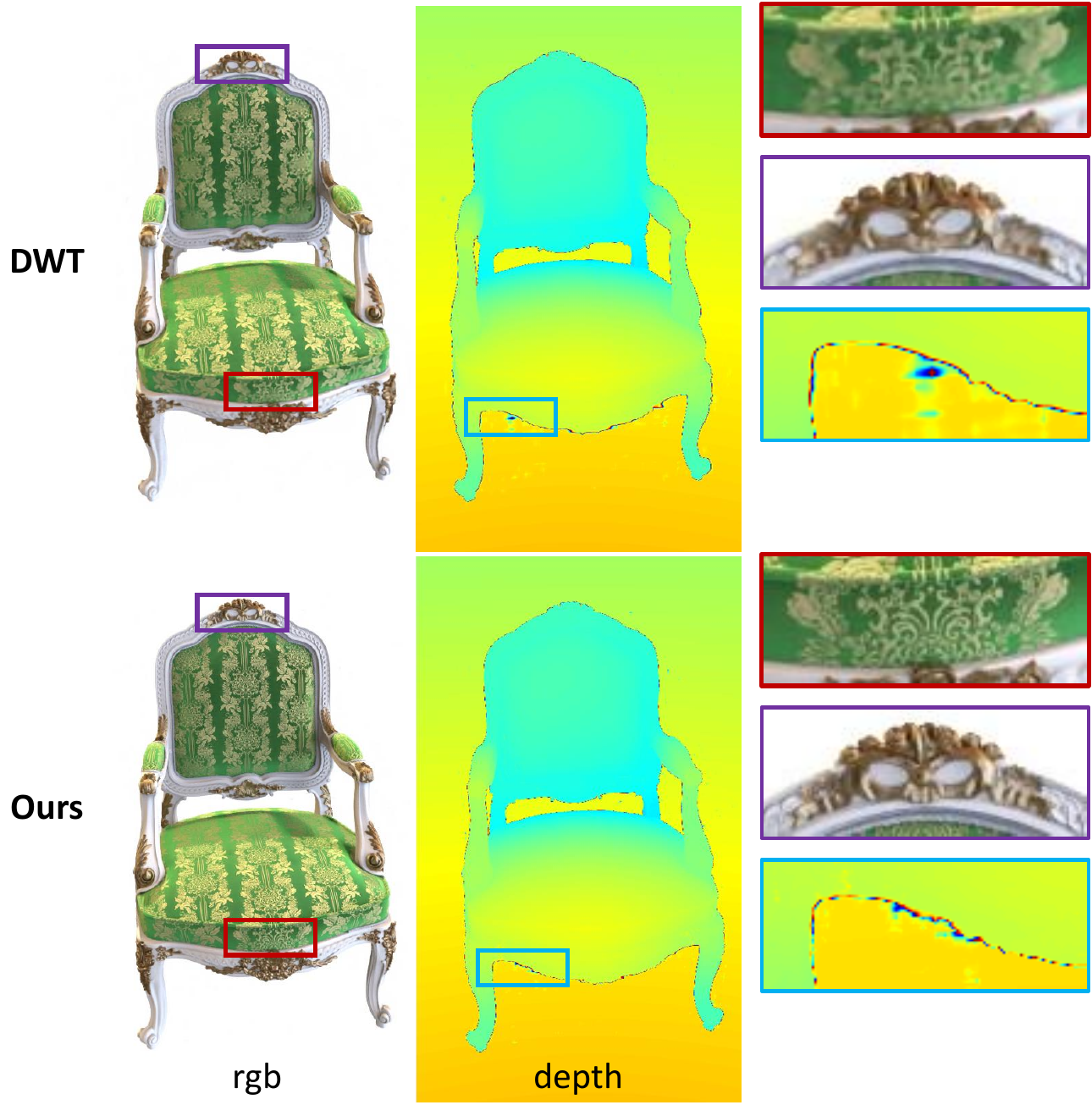}
    \caption{
    \textbf{Visual Comparison of Static Scenes on NSVF Data.}
    Two representative patches are selected for closer inspection. Our method, free from the DWT limitations of shift variance and direction ambiguity, achieves superior texture reconstruction performance.}
    \label{fig:static_comp}
\vspace{-1.5em}
\end{figure}

Figure \ref{fig:static_comp} highlights the qualitative differences between DWT-based solutions and our proposed direction-aware method. In static scenes, our solution excels in reconstructing texture details compared to DWT representation, which is less sensitive to lines and edges patterns due to shift variance and direction ambiguity.

\subsection{Ablations}
\noindent
\textbf{Wavelet Function.} We analyze the impact of different wavelet functions on reconstruction quality, aiming to facilitate a comparison between our direction-aware representation and DWT wavelet. The evaluation is conducted on NSVF data \cite{liu2020neural}, where several complex wavelet functions with the approximate half-sample delay property—Antonini, LeGall, and two Near Symmetric filter banks (Near Symmetric A and Near Symmetric B)—are selected for comparison. Table \ref{tab:wavelet_function} reveals that the choice of different wavelets has minimal effect on reconstruction quality. Even the worst-performing wavelet function outperforms the discrete wavelet transform, underscoring the advantages of our direction-aware representation.
\begin{table}[t]
    \centering
    \caption{
    \textbf{Impact of Wavelet Transform Type/Function}, on reconstruction performance, evaluated on NSVF data..
    }
    \vspace{-.5em}
    \begin{tabular}{ccc}
    \toprule
        Wavelet Type&Wavelet Function & PSNR $\uparrow$ \\
        \midrule
        \multirow{4}*{DWT}& Haar&34.61\\
        ~& Coiflets 1 &34.56\\
        ~& \textbf{biorthogonal 4.4}&\textbf{34.67}\\
        ~& Daubechies 4 &34.44\\
        \midrule
         \multirow{4}*{DTCWT}&Antonini & 36.10\\
         ~&LeGall & 36.14\\
         ~&\textbf{Near Symmetric A}& \textbf{36.24}\\
         ~&Near Symmetric B& 36.17\\
         \bottomrule
    \end{tabular}
    \label{tab:wavelet_function}
\end{table}

\noindent
\textbf{Sparsity Analysis}. We evaluate the sparsity of our direction-aware representation by varying the sparsity level using different $\lambda_m$ values on the NSVF dataset. As depicted in Table \ref{tab:sparsity}, our direction-aware representation consistently achieves over 99\% sparsity. This remarkable sparsity, coupled with a model size of approximately 1MB, demonstrates the efficiency of our method in modeling static scenes 
while outperforming
state-of-the-art sparse representation methods.

\begin{table}[t]
    \centering
    \caption{
    \textbf{Sparsity Analysis of Direction-Aware Representation}, evaluated on NVSF data.}
    \vspace{-.5em}
    \resizebox{\linewidth}{!}{
    \begin{tabular}{cccc}
    \toprule
         $\lambda_m$&Sparsity $\uparrow$&Model Size (MB)  $\downarrow$ &PSNR $\uparrow$ \\
         \midrule
         $1.0\times 10^{-10}$&\textbf{99.2\%}& \textbf{1.16 MB}&35.36\\
         $5.0\times 10^{-11}$&97.3\%& 3.18 MB&35.81\\
         $2.5\times 10^{-11}$&94.2\%& 8.98 MB&\textbf{36.24}\\
         $0$&-&135 MB&36.34\\
         \bottomrule
    \end{tabular}
    }
    \label{tab:sparsity}
\vspace{-1.25em}
\end{table}

\noindent
\textbf{Wavelet Levels.} We investigated the impact of scene reconstruction performance across different wavelet levels, and the results are presented in supplementary material. We observed that increasing the wavelet level did not lead to significant performance improvements. Conversely, we noted a substantial increase in both training time and model size with the increment of wavelet level. As a result, throughout all experiments, we consistently set the wavelet level to 1.

%% file: 10_conclusion.tex
\section{Discussion}
\label{sec:conclusion}

\noindent
\textbf{Limitations.} 
Our method is limited for scenarios with extremely sparse observations, as seen in D-NeRF-like datasets. DaReNeRF does not incorporate a deformation field into the model, lacking a robust information-sharing mechanism to learn 3D structures from very sparse views. Another limitation of our proposed direction-aware representation is its lower compactness compared to DWT representation, preventing DaReNeRF from achieving extremely small model sizes, such as less than 1MB on static scene. Exploring more compact methods to construct direction-aware representations would be an interesting direction for future research.

\noindent
\textbf{Conclusion.} We introduced a novel direction-aware representation capable of effectively capturing information from six different directions. The shift-invariant and direction-selective nature of our proposed representation enables the high-fidelity reconstruction of challenging dynamic scenes without the need for prior knowledge about the scene dynamics. Despite introducing some storage redundancy, we mitigate this by incorporating trainable masks for both static and dynamic scenes, resulting in a model size comparable to recent methods. We believe that this simple yet effective representation has the potential to simplify and streamline dynamic NeRFs, providing a more accessible and efficient solution for complex scene modeling.

%% file: 12_appendix.tex
In this supplementary material, we provide further methodological context and implementation details to facilitate reproducibility of our framework DaReNeRF. We also showcase additional quantitative and qualitative results to further highlight the contributions claimed in the paper.

\section{Dual-Tree Complex Wavelet Transform}
\begin{figure}[!h]
    \centering
    \includegraphics[width=1.\linewidth]{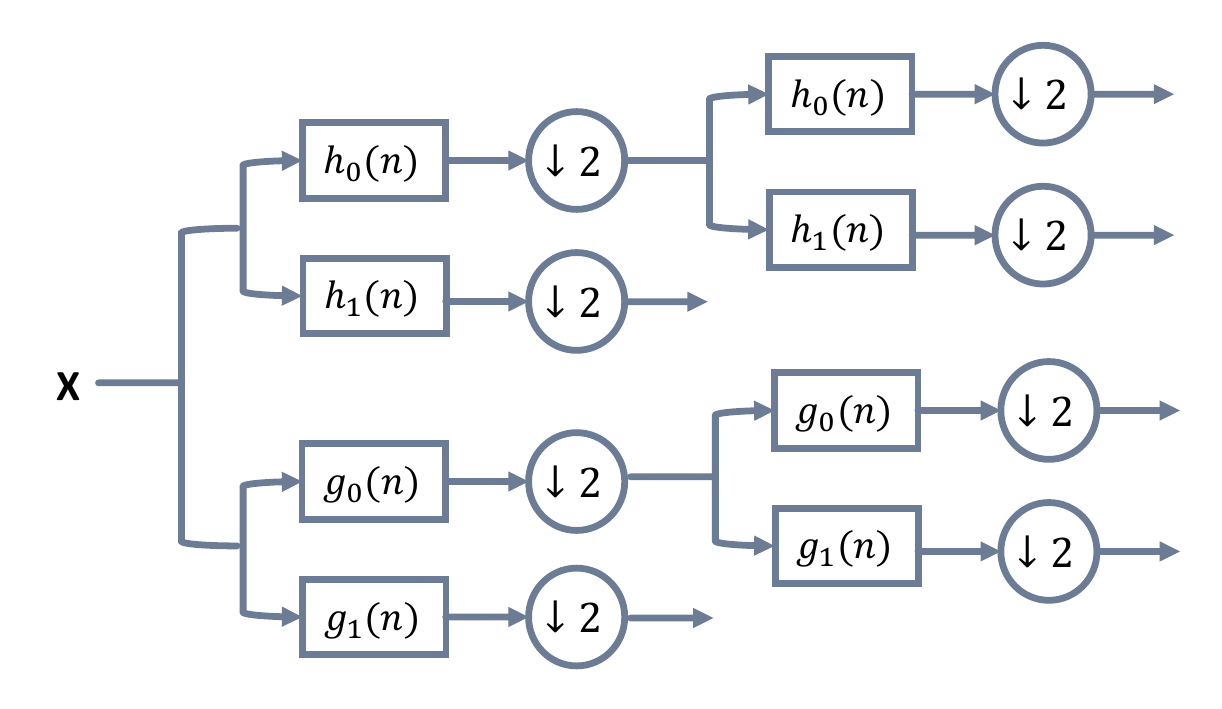}
    \caption{\textbf{Analysis Filter Bank}, for the dual tree complex wavelet transfrom.}
        \label{fig:DTCWT_filter_bank}
\end{figure}

The idea of dual-tree complex wavelet transform (DTCWT) \cite{selesnick2005dual} is quite straightforward. The DTCWT employs two real discrete wavelet transforms (DWTs). The first DWT gives the real part of the transform while the second DWT gives the imaginary part. The analysis filter banks used to implement the DTCWT is illustrated in Figure \ref{fig:DTCWT_filter_bank}. Here $h_0(n)$, $h_1(n)$ denote the low-pass/high-pass filter pair for upper filter bank, and $g_0(n)$, $g_1(n)$ denote the low-pass/high-pass filter pair for the lower filter bank. The two real wavelets associated with each of the two real wavelet transforms as $\psi_h(t)$ and $\psi_g(t)$. And the complex wavelet can be denoted as $\psi(t)=\psi_h(t)+j\psi_g(t)$. The $\psi_g(t)$ is approximately the Hilbert transform of $\psi_h(t)$. The 2D DTCWT $\psi(x,y) = \psi(x)\psi(y)$ associated with the row-column implementation of the wavelet transform, where $\psi(x)$ is a complex wavelet given by $\psi(x)=\psi_h(x)+j\psi_g(x)$. Then we obtain for $\psi(x,y)$ the expression:
\begin{small}
\begin{equation}
\begin{split}
    \psi(x,y)&=[\psi_h(x)+j\psi_g(x)][\psi_h(y)+j\psi_g(y)] \\
    &= \psi_h(x)\psi_h(y)-\psi_g(x)\psi_g(y) \\
    &+ j[\psi_g(x)\psi_h(y)+\psi_h(x)\psi_g(y)]
\end{split}
\label{euqation_1}
\end{equation}
\end{small}
The spectrum of $\psi_h(x)\psi_h(y)-\psi_g(x)\psi_g(y)$ which corresponds to the real part of $\psi(x,y)$ is supported in two quadrants of the 2D frequency plane and is oriented at $-45^\circ$. Note that the $\psi_h(x)\psi_h(y)$ is the HH wavelet of a separable 2D real wavelet transform implemented using the filter pair $\{h_0(n),h_1(n)\}$. Similarly, $\psi_g(x)\psi_g(y)$ is the HH wavelet of a real separable wavelet transform, implemented using the filters $\{g_0(n),g_1(n)\}$. To obtain a real 2D wavelet oriented at $+45^\circ$, we consider now the complex 2-D wavelet $\psi(x,y)=\psi(x)\overline{\psi(y)}$, where $\overline{\psi(y)}$ represents the complex conjugate of $\psi(y)$. This gives us the following expression:
\begin{small}
\begin{equation}
\begin{split}
    \psi(x,y)&=[\psi_h(x)+j\psi_g(x)][\overline{\psi_h(y)+j\psi_g(y)}] \\
    &= \psi_h(x)\psi_h(y)+\psi_g(x)\psi_g(y) \\
    &+ j[\psi_g(x)\psi_h(y)-\psi_h(x)\psi_g(y)]
\end{split}
\label{euqation_2}
\end{equation}
\end{small}
The spectrum of $\psi_h(x)\psi_h(y)+\psi_g(x)\psi_g(y)$ is supported in two quadrants of the 2D frequency plane and is oriented at $+45^\circ$. We could obtain four more oriented real 2D wavelets by repeating the above procedure on the following complex 2-D wavelets: $\phi(x)\psi(y)$, $\psi(x)\phi(y)$, $\phi(x)\overline{\psi(y)}$ and $\psi(x)\overline{\phi(y)}$, where $\psi(x) = \psi_h(x)+j\psi_g(y)$ and $\phi(x)=\phi_h(x)+j\phi_g(y)$. By taking the real part of each of these four complex wavelets, we obtain four real oriented 2D wavelets, in additional to the two already obtain in \ref{euqation_1} and \ref{euqation_2}:
\begin{small}
\begin{equation}
    \psi_i(x,y) = \dfrac{1}{\sqrt{2}}(\psi_{1,i}(x,y)-\psi_{2,i}(x,y)),
\label{euqation_3}
\end{equation}
\end{small}
\begin{small}
\begin{equation}
    \psi_{i+3}(x,y) = \dfrac{1}{\sqrt{2}}(\psi_{1,i}(x,y)+\psi_{2,i}(x,y))
\label{euqation_4}
\end{equation}
\end{small}
for $i=1,2,3$, where the two separable 2-D wavelet bases are defined in the usual manner:
\begin{small}
\begin{equation}
\begin{split}
    \psi_{1,1}(x,y)&=\phi_h(x)\psi_h(y), \psi_{2,1}(x,y)=\phi_g(x)\psi_g(y), \\
    \psi_{1,2}(x,y)&=\psi_h(x)\phi_h(y), \psi_{2,2}(x,y)=\psi_g(x)\phi_g(y), \\
    \psi_{1,3}(x,y)&=\psi_h(x)\psi_h(y), \psi_{2,3}(x,y)=\psi_g(x)\psi_g(y),
\end{split}
\end{equation}
\end{small}
We have used the normalization $\dfrac{1}{\sqrt{2}}$ only so that the sum and difference operation constitutes an orthonormal operation. From the imaginary parts of $\psi(x)\psi(y)$, $\psi(x)\overline{\psi(y)}$, $\phi(x)\psi(y)$, $\psi(x)\phi(y)$, $\phi(x)\overline{\psi(y)}$ and $\psi(x)\overline{\phi(y)}$, we could obtain six oriented wavelets given by:
\begin{small}
\begin{equation}
    \psi_i(x,y) = \dfrac{1}{\sqrt{2}}(\psi_{3,i}(x,y)+\psi_{4,i}(x,y)),
\end{equation}
\end{small}
\begin{small}
\begin{equation}
    \psi_{i+3}(x,y) = \dfrac{1}{\sqrt{2}}(\psi_{3,i}(x,y)-\psi_{4,i}(x,y))
\end{equation}
\end{small}
for $i=1,2,3$, where the two separable 2D wavelet bases are defined as:
\begin{small}
\begin{equation}
\begin{split}
    \psi_{3,1}(x,y)&=\phi_g(x)\psi_h(y), \psi_{4,1}(x,y)=\phi_h(x)\psi_g(y), \\
    \psi_{3,2}(x,y)&=\psi_g(x)\phi_h(y), \psi_{4,2}(x,y)=\psi_h(x)\phi_g(y), \\
    \psi_{3,3}(x,y)&=\psi_g(x)\psi_h(y), \psi_{4,3}(x,y)=\psi_h(x)\psi_g(y),
\end{split}
\end{equation}
\end{small}
Thus we could obtain six oriented wavelets from both real and imaginary part. 

\section{Additional Results on Various Datasets}

\subsection{Plenoptic Video Dataset \cite{li2022neural}}
\begin{figure*}[ht]
    \centering
    \includegraphics[width=\linewidth]{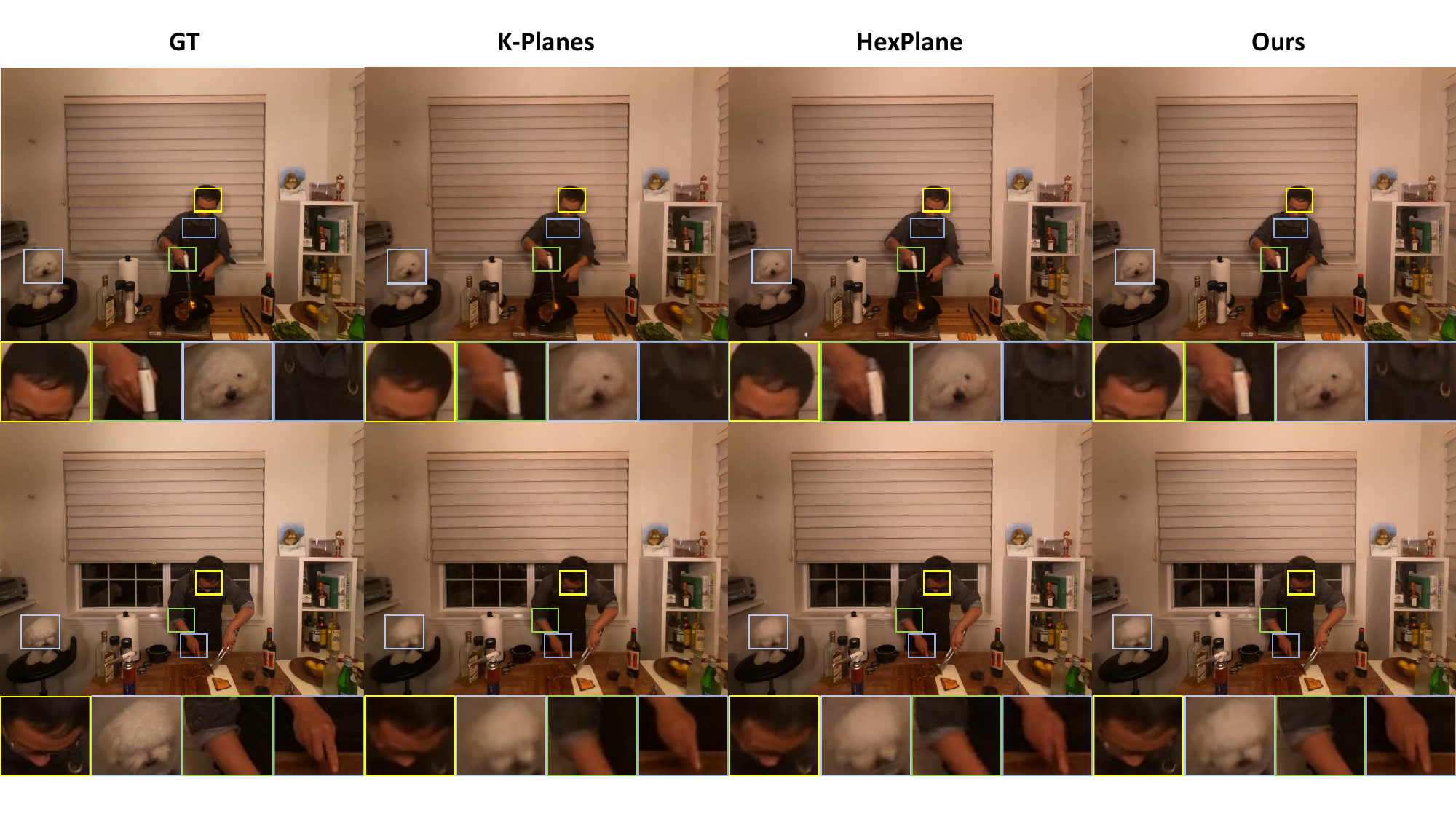}
    \caption{Visual Comparison on Dynamic Scenes (Plenoptic Data). K-Planes and HexPlane are concurrent decomposition-based methods.
As shown in the four zoomed-in patches, our method better reconstructs fine details and captures motion.}
    \label{fig:vis_comparisons}
\end{figure*}

\begin{figure*}[ht]
    \centering
    \includegraphics[width=\linewidth,height=\textheight]{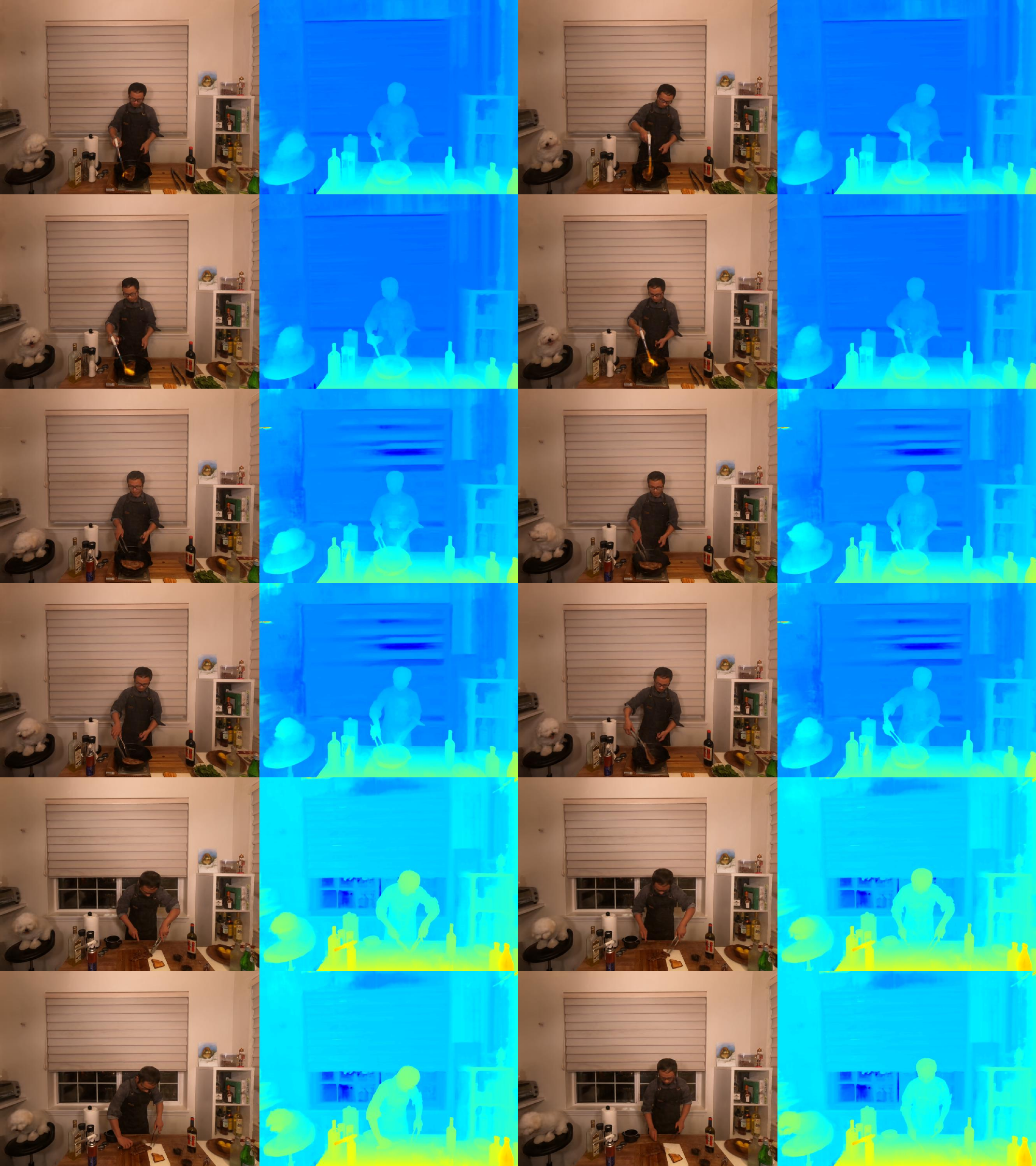}
    \caption{Visualizations on \texttt{flame steak}, \texttt{sear steak} and \texttt{cut roasted beef} scene.}
    \label{fig:fig_1_plenoptic}
\end{figure*}

\begin{figure*}[ht]
    \centering
    \includegraphics[width=\linewidth,height=\textheight]{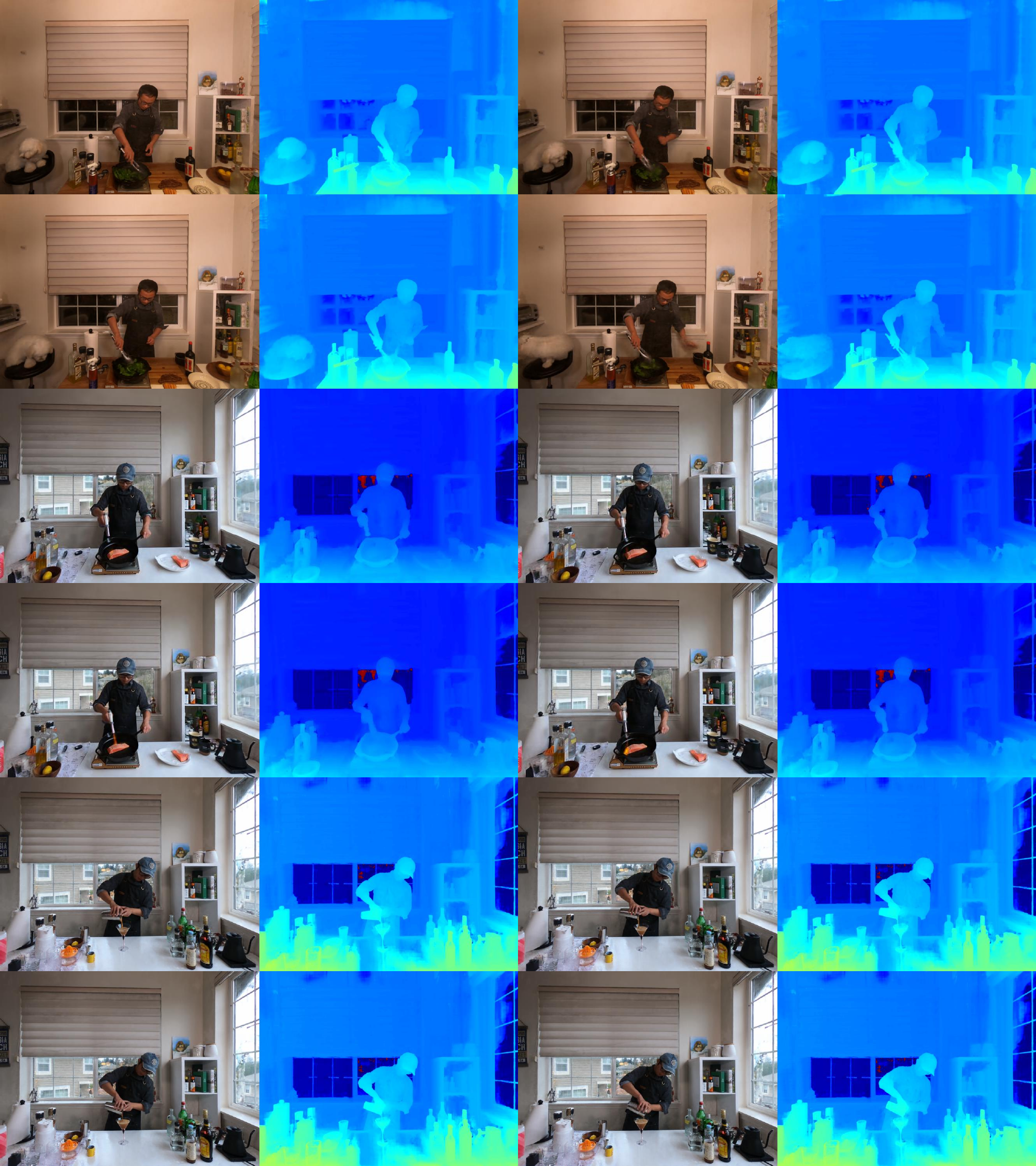}
    \caption{Visualizations on \texttt{cook spinach}, \texttt{flame salmon} and \texttt{coffee martini} scene.}
    \label{fig:fig_2_plenoptic}
\end{figure*}

The quantitative results for each scene are presented in Table \ref{tab:plenoptic_quantitative}, while additional visualizations comparing DaReNeRF with current state-of-the-art methods, HexPlane \cite{cao2023hexplane} and K-Planes \cite{fridovich2023k}, are provided in Figure \ref{fig:vis_comparisons}. Notably, DaReNeRF demonstrates superior recovery of texture details. Furthermore, comprehensive visualizations of DaReNeRF on all six scenes in the Plenoptic dataset are shown in Figure \ref{fig:fig_1_plenoptic} and Figure \ref{fig:fig_2_plenoptic}.

\begin{table*}[ht]
    \centering
    \caption{Results of Plenoptic Video Dataset. We report results of each scene}
    \resizebox{\linewidth}{!}{
    \begin{tabular}{cccccccccc}
    \toprule
         \multirow{2}*{Model}&\multicolumn{3}{c}{Flame Salmon}&\multicolumn{3}{c}{Cook Spinach}&\multicolumn{3}{c}{Cut Roasted Beef} \\
         ~&PSNR $\uparrow$ &D-SSIM $\downarrow$ &LPIPS $\downarrow$&PSNR $\uparrow$ &D-SSIM $\downarrow$ &LPIPS $\downarrow$&PSNR $\uparrow$ &D-SSIM $\downarrow$ &LPIPS $\downarrow$\\
    \midrule
    DaReNeRF-S&30.294&0.015&0.089&32.630&0.013&0.100&33.087&0.013&0.092\\
    \textbf{DaReNeRF}&\textbf{30.441}&\textbf{0.012}&\textbf{0.084}&\textbf{32.836}&\textbf{0.011}&\textbf{0.090}&\textbf{33.200}&\textbf{0.011}&\textbf{0.091}\\
    \midrule
    \multirow{2}*{ }&\multicolumn{3}{c}{Flame Steak}&\multicolumn{3}{c}{Sear Steak}&\multicolumn{3}{c}{Coffee Martini} \\
    DaReNeRF-S&33.259&0.011&0.081&33.179&0.011&0.075&30.160&0.016&0.092\\
    \textbf{DaReNeRF}&\textbf{33.524}&\textbf{0.009}&\textbf{0.077}&\textbf{33.351}&\textbf{0.009}&\textbf{0.072}&\textbf{30.193}&\textbf{0.014}&\textbf{0.089}\\
    \bottomrule
    \end{tabular}
    }
    \label{tab:plenoptic_quantitative}
\end{table*}

\subsection{D-NeRF Dataset \cite{pumarola2021d}}
\begin{figure*}[ht]
    \centering
    \includegraphics[width=\linewidth]{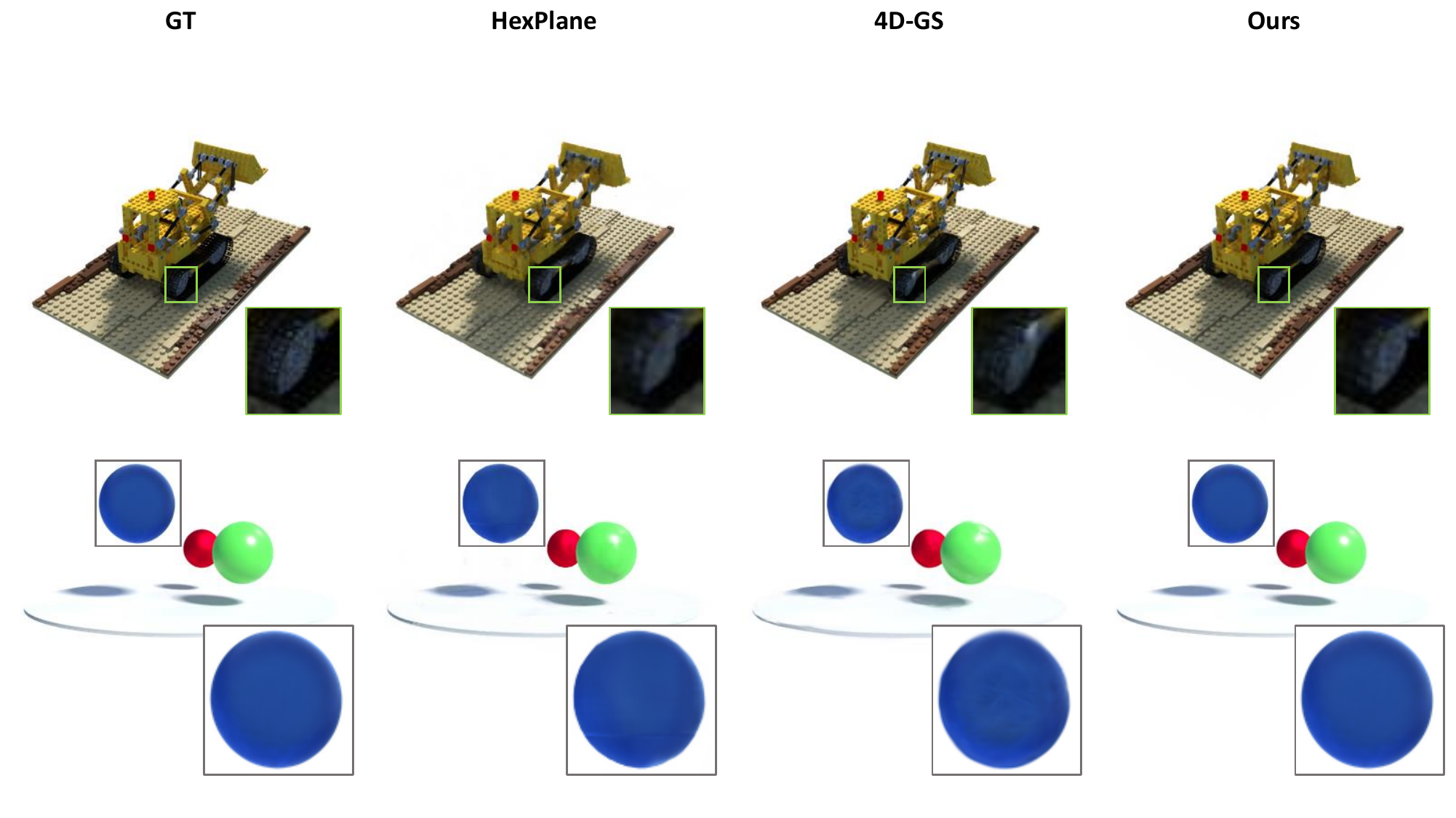}
    \caption{Visual Comparison on Dynamic Scenes (D-NeRF Data). 4D-GS and HexPlane are decomposition-based and deformation-based methods.}
    \label{fig:dnerf_compare}
\end{figure*}

\begin{figure*}[ht]
    \centering
    \includegraphics[width=\linewidth,height=\textheight]{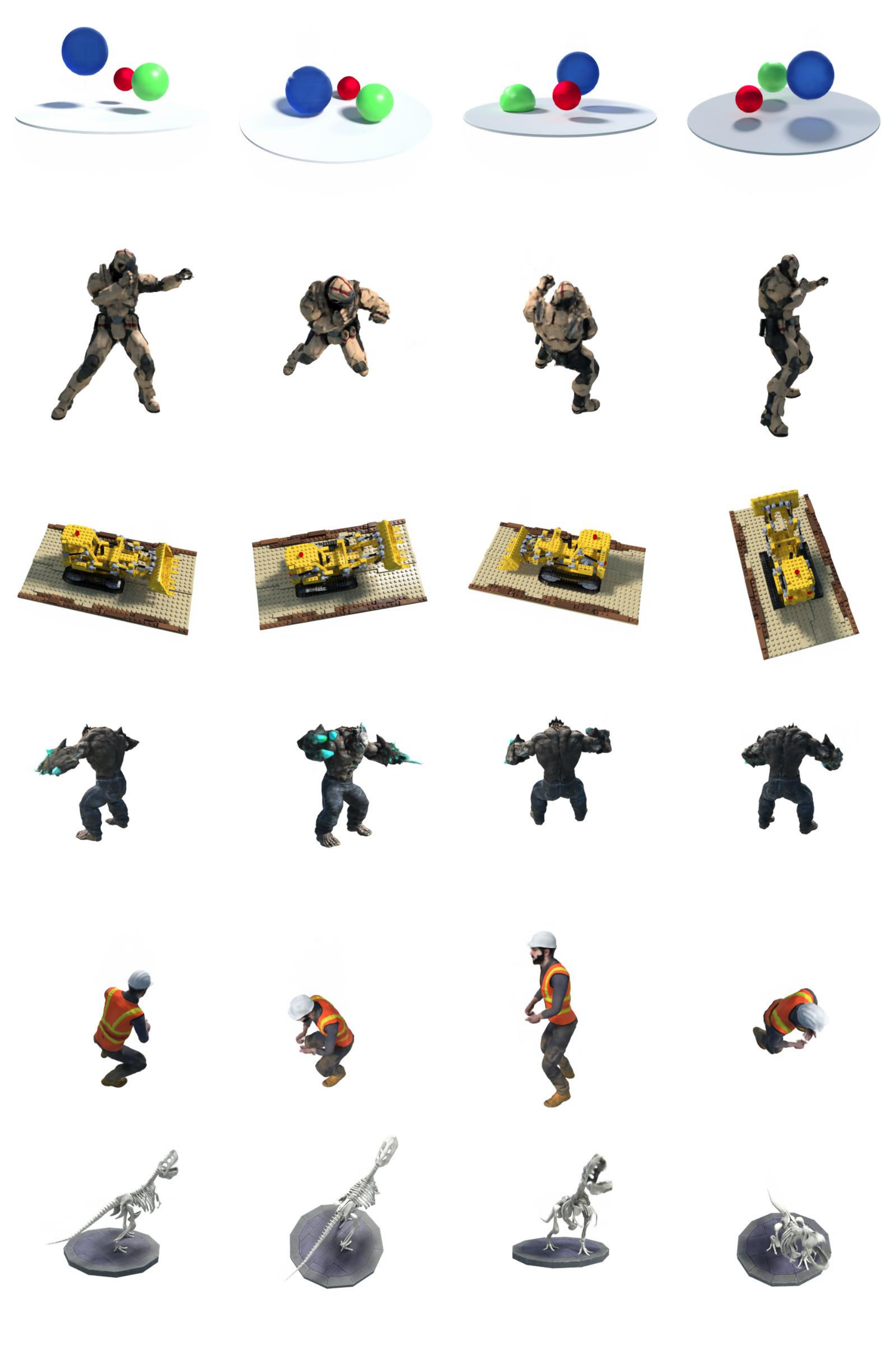}
    \caption{Visualizations on D-NeRF dataset}
    \label{fig:dnerf_good}
\end{figure*}

\begin{figure*}[ht]
    \centering
    \includegraphics[width=\linewidth]{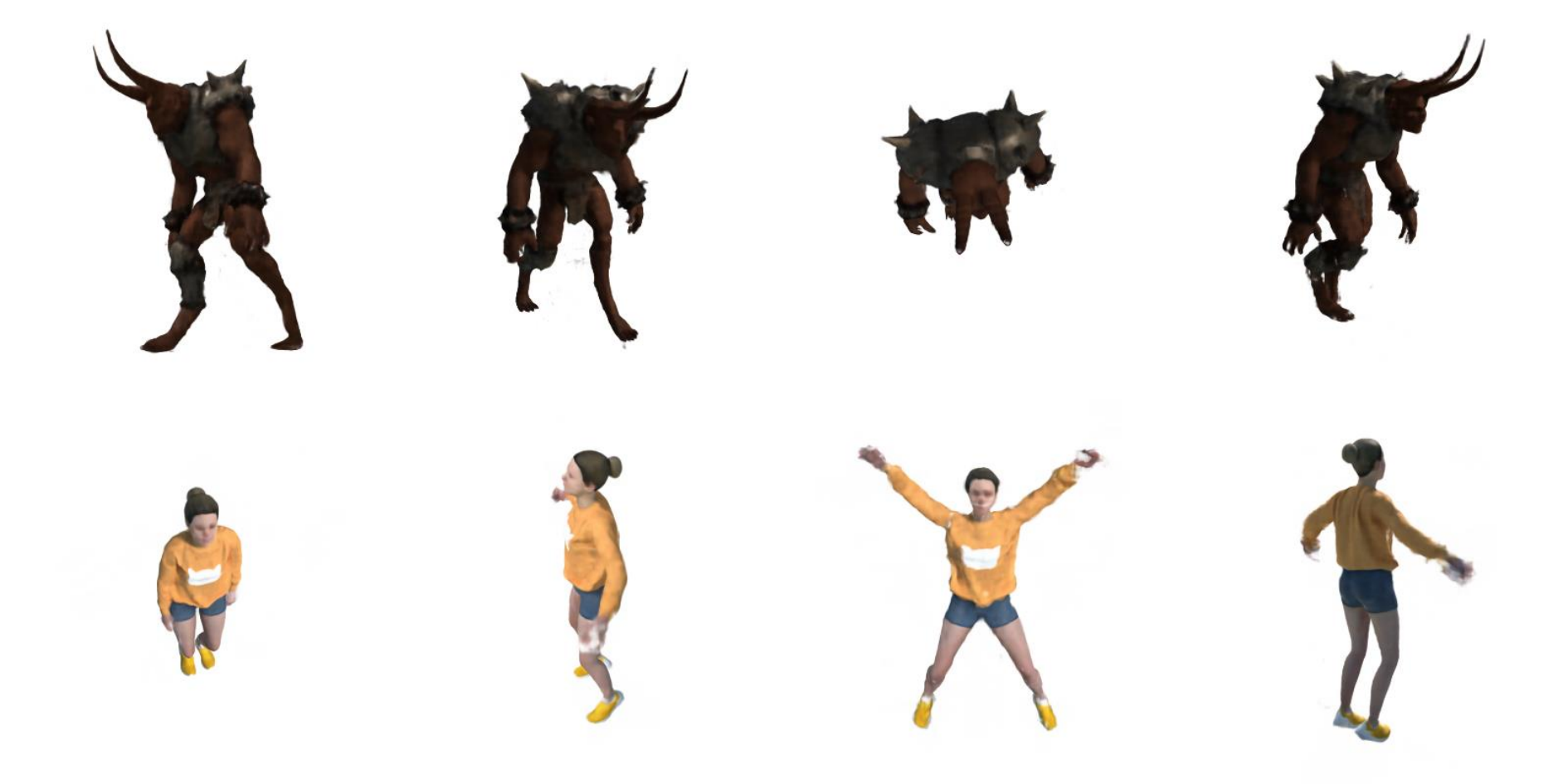}
    \caption{Visualizations on failure cases from D-NeRF dataset}
    \label{fig:dnerf_fail}
\end{figure*}

We provide quantitative results for each scene in Table \ref{tab:d_nerf_quan}, while additional visualizations comparing DaReNeRF with current state-of-the-art methods, HexPlane \cite{cao2023hexplane} and 4D-GS \cite{wu20234d}, are shared in Figure \ref{fig:dnerf_compare}. We also provide further visualization in a video attached to this supplementary material. 
Remarkably, although 4D-GS incorporates a deformation field, DaReNeRF still outperforms it in certain cases from the D-NeRF dataset. Furthermore, comprehensive visualizations of DaReNeRF on six scenes in the Plenoptic dataset are shown in Figure \ref{fig:dnerf_good} and the failure cases are shown in Figure \ref{fig:dnerf_fail}.

\begin{table*}[ht]
    \centering
    \caption{Results of D-NeRF Dataset. We report results of each scene}
    \resizebox{\linewidth}{!}{
    \begin{tabular}{cccccccccc}
    \toprule
         \multirow{2}*{Model}&\multicolumn{3}{c}{Hell Warrior}&\multicolumn{3}{c}{Mutant}&\multicolumn{3}{c}{Hook} \\
         ~&PSNR $\uparrow$ &SSIM $\uparrow$ &LPIPS $\downarrow$&PSNR $\uparrow$ &SSIM $\uparrow$ &LPIPS $\downarrow$&PSNR $\uparrow$ &SSIM $\uparrow$ &LPIPS $\downarrow$\\
         \midrule
         T-NeRF&23.19&0.93&0.08&30.56&0.96&0.04&27.21&0.94&0.06  \\
         D-NeRF&25.02&0.95&0.06&31.29&0.97&0.02&29.25&0.96&0.11 \\
         TiNeuVox-S&27.00&0.95&0.09&31.09&0.96&0.05&29.30&0.95&0.07\\
         TiNeuVox-B&28.17&0.97&0.07&33.61&0.98&0.03&31.45&0.97&0.05\\
         HexPlane&24.24&0.94&0.07&33.79&0.98&0.03&28.71&0.96&0.05\\
         \midrule
         DaReNeRF-S&25.71&0.95&0.04&34.08&0.98&0.02&29.04&0.96&0.04\\
         DaReNeRF&25.82&0.95&0.04&34.17&0.98&0.01&28.96&0.96&0.04\\
         \midrule
         \multirow{2}*{ }&\multicolumn{3}{c}{Bouncing Balls}&\multicolumn{3}{c}{Lego}&\multicolumn{3}{c}{T-Rex} \\
         ~&PSNR $\uparrow$ &SSIM $\uparrow$ &LPIPS $\downarrow$&PSNR $\uparrow$ &SSIM $\uparrow$ &LPIPS $\downarrow$&PSNR $\uparrow$ &SSIM $\uparrow$ &LPIPS $\downarrow$\\
         \midrule
         T-NeRF&37.81&0.98&0.12&23.82&0.90&0.15&30.19&0.96&0.13\\
         D-NeRF&38.93&0.98&0.10&21.64&0.83&0.16&31.75&0.97&0.03\\
         TiNeuVox-S&39.05&0.99&0.06&24.35&0.88&0.13&29.95&0.96&0.06\\
         TiNeuVox-B&40.73&0.99&0.04&25.02&0.92&0.07&32.70&0.98&0.03\\
         HexPlane&39.69&0.99&0.03&25.22&0.94&0.04&30.67&0.98&0.03\\
         \midrule
         DaReNeRF-S&42.24&0.99&0.01&25.24&0.94&0.03&31.75&0.98&0.03\\
         DaReNeRF&42.26&0.99&0.01&25.44&0.95&0.03&32.21&0.98&0.02\\
         \midrule
         \multirow{2}*{ }&\multicolumn{3}{c}{Stand Up}&\multicolumn{3}{c}{Jumping Jacks}&\multicolumn{3}{c}{Average} \\
         ~&PSNR $\uparrow$ &SSIM $\uparrow$ &LPIPS $\downarrow$&PSNR $\uparrow$ &SSIM $\uparrow$ &LPIPS $\downarrow$&PSNR $\uparrow$ &SSIM $\uparrow$ &LPIPS $\downarrow$\\
         \midrule
         T-NeRF&31.24&0.97&0.02&32.01&0.97&0.03&29.51&0.95&0.08\\
         D-NeRF&32.79&0.98&0.02&32.80&0.98&0.03&30.50&0.95&0.07\\
         TiNeuVox-S&32.89&0.98&0.03&32.33&0.97&0.04&30.75&0.96&0.07\\
         TiNeuVox-B&35.43&0.99&0.02&34.23&0.98&0.03&32.64&0.97&0.04\\
         HexPlane&34.36&0.98&0.02&31.65&0.97&0.04&31.04&0.94&0.04\\
         \midrule
         DaReNeRF-S&34.47&0.98&0.02&31.99&0.97&0.03&31.82&0.97&0.03\\
         DaReNeRF&34.58&0.98&0.02&32.21&0.97&0.03&31.95&0.97&0.03\\
         \bottomrule

    \end{tabular}
    }
    \label{tab:d_nerf_quan}
\end{table*}

\subsection{NeRF Synthetic Dataset}
\begin{figure*}
    \centering
    \includegraphics[width=\linewidth]{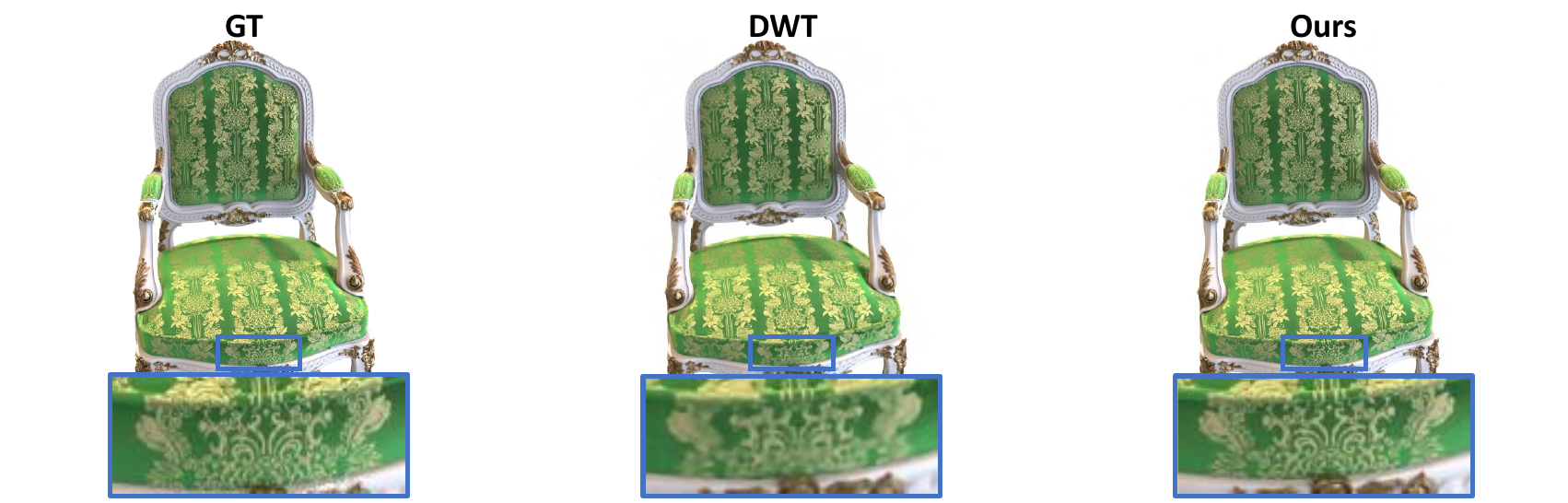}
    \caption{Visual comparison on NeRF synthetic dataset.}
    \label{fig:nerf_synthetic_1}
\end{figure*}

\begin{figure*}
    \centering
    \includegraphics[width=0.7\linewidth,height=\textheight]{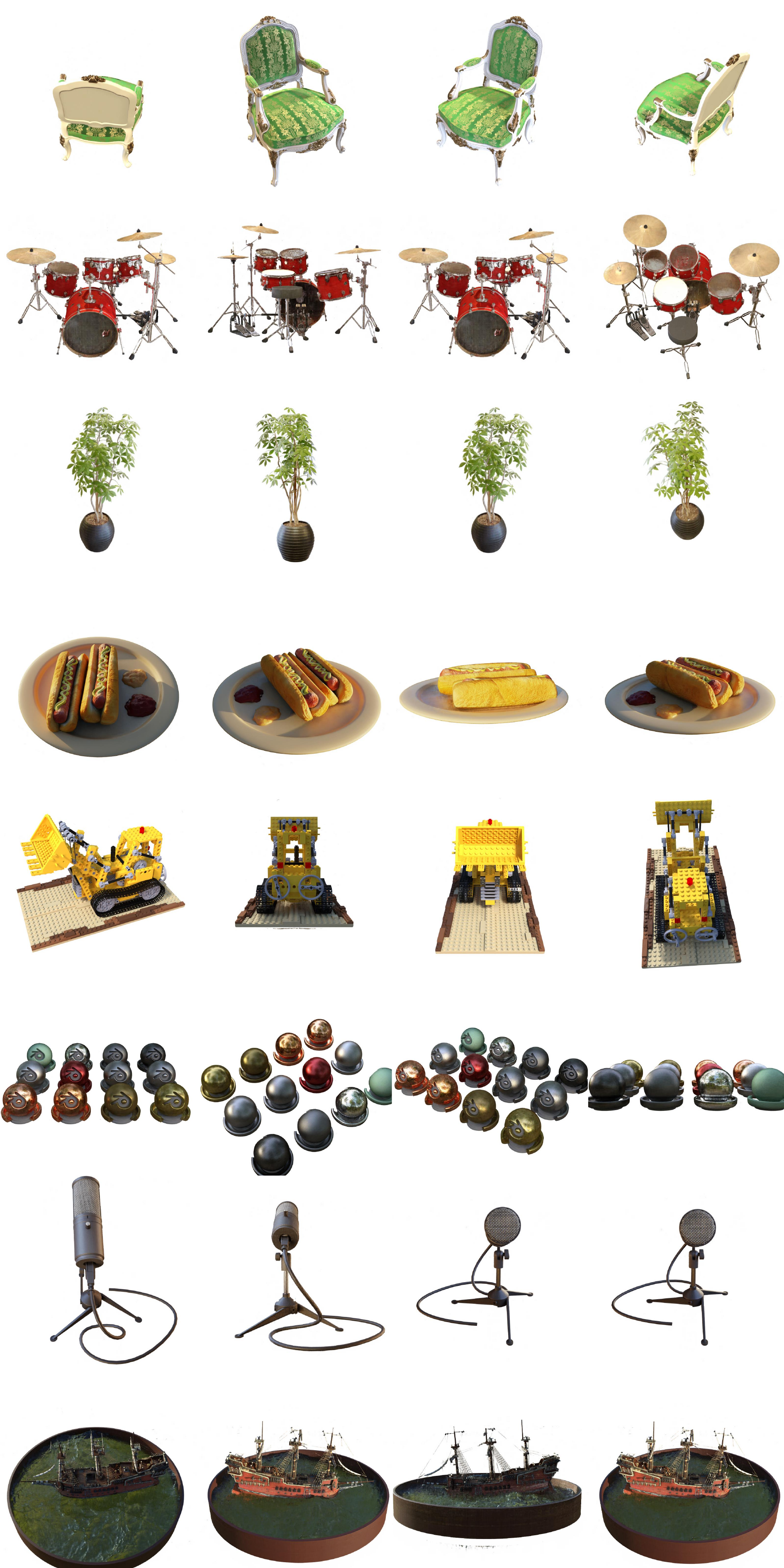}
    \caption{Visualizations on NeRF synthetic dataset.}
    \label{fig:nerf_synthetic_2}
\end{figure*}

The quantitative results for each case are presented in Table \ref{tab:quant_nerf}, while additional visualizations comparing our representation with DWT \cite{rho2023masked} based representation method, are shown in Figure \ref{fig:nerf_synthetic_1}.  Furthermore, comprehensive visualizations of eight scenes in the NeRF dataset are shown in Figure \ref{fig:nerf_synthetic_2} and in the attached video.

\begin{table*}[ht]
    \centering
    \caption{Results of NeRF Synthetic Dataset}
    \resizebox{\linewidth}{!}{
    \begin{tabular}{cccccccccccc}
    \toprule
         Bit Precision&Method&Size(MB)&Avg&Chair&Drums&Ficus&Hotdog&Lego&Materials&Mic&Ship  \\
         \midrule
         32-bit&KiloNeRF&$\leq$ 100&31.00&32.91&25.25&29.76&35.56&33.02&29.20&33.06&29.23\\
         32-bit&CCNeRF (CP)&4.4&30.55&-&-&-&-&-&-&-&-\\
         8-bit$^\ast$&NeRF&1.25&31.52&33.82&24.94&30.33&36.70&32.96&29.77&34.41&29.25\\
         8-bit&cNeRF&0.70&30.49&32.28&24.85&30.58&34.95&31.98&29.17&32.21&28.24\\
         8-bit$^\ast$&PREF&9.88&31.56&34.55&25.15&32.17&35.73&34.59&29.09&32.64&28.58\\
         8-bit$^\ast$&VM-192&17.93&32.91&35.64&25.98&33.57&37.26&36.04&29.87&34.33&30.64\\
         8-bit$^\ast$&VM-192 (300) + DWT&0.83&31.95&34.14&25.53&32.87&36.08&34.93&29.42&33.48&29.15\\
         \midrule
         8-bit$^\ast$&VM-192 (300) + Ours&8.91&32.42&36.05&29.40&35.26&36.37&25.58&33.26&29.82&33.63\\
         \bottomrule
         & 
    \end{tabular}
    }
    
    \label{tab:quant_nerf}
\end{table*}

\subsection{NSVF Synthetic Dataset}
\begin{figure*}
    \centering
    \includegraphics[width=\linewidth]{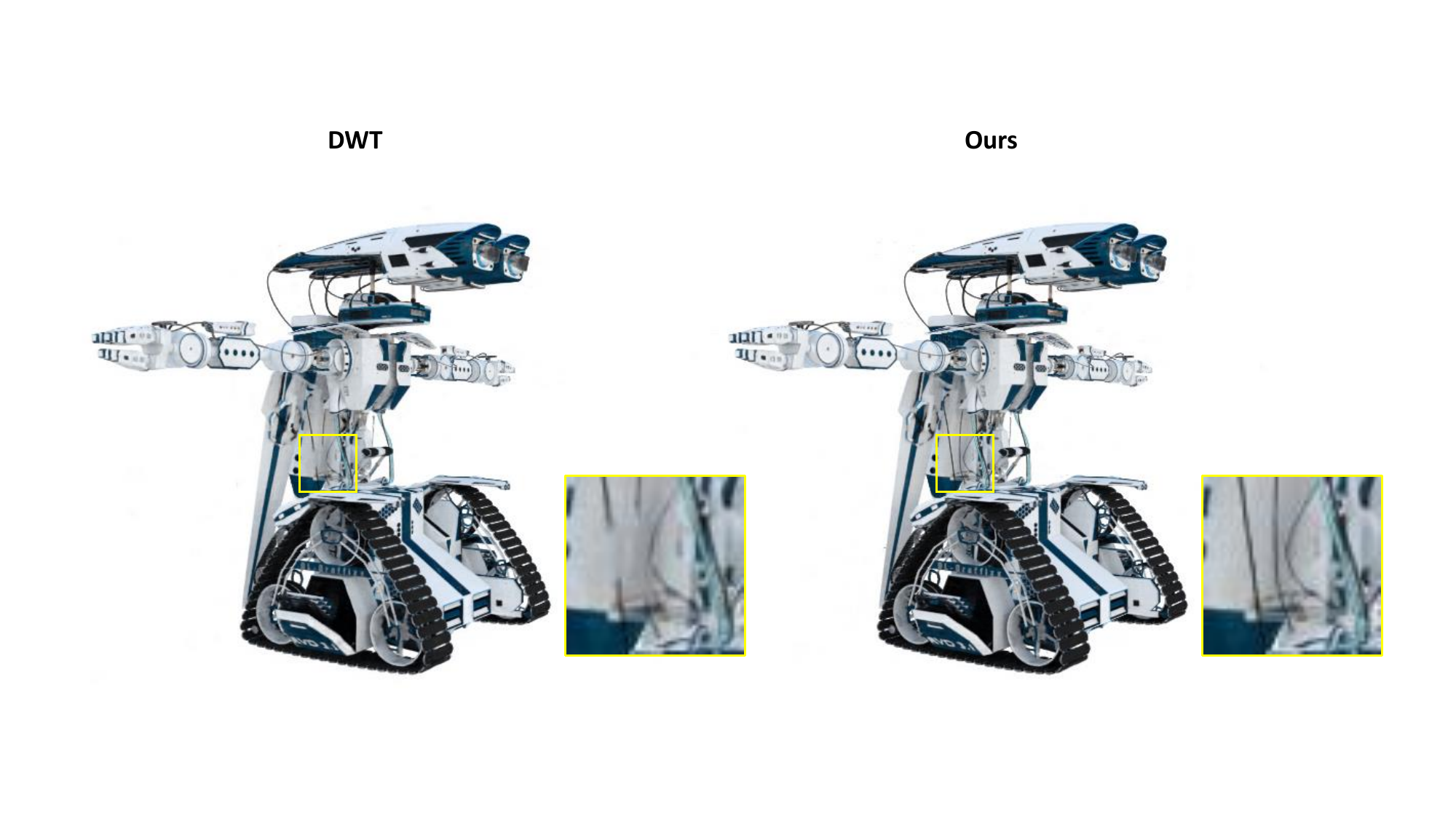}
    \caption{Visual comparison on NSVF synthetic dataset.}
    \label{fig:nsvf_synthetic_1}
\end{figure*}

\begin{figure*}
    \centering
    \includegraphics[width=0.7\linewidth,height=\textheight]{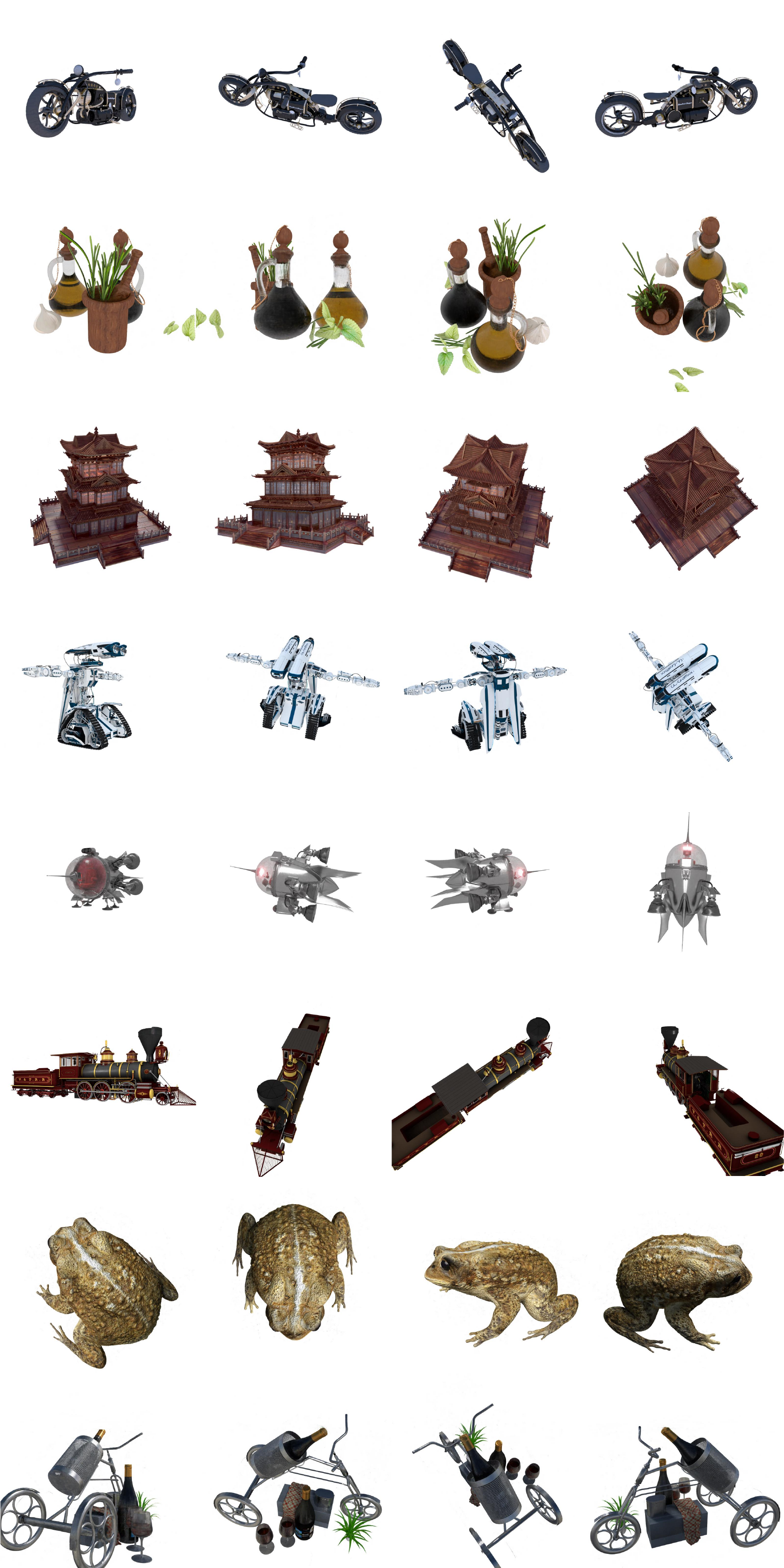}
    \caption{Visualizations on NSVF synthetic dataset.}
    \label{fig:nsvf_synthetic_2}
\end{figure*}

The quantitative results for each case are presented in Table \ref{tab:quant_nsvf}, while additional visualizations comparing our representation with DWT \cite{rho2023masked} based representation method, are shown in Figure \ref{fig:nsvf_synthetic_1}.  Furthermore, comprehensive visualizations of eight scenes in the NSVF dataset are shown in Figure \ref{fig:nsvf_synthetic_2}.

\begin{table*}[ht]
    \centering
    \caption{Results of NSVF Synthetic Dataset}
    \resizebox{\linewidth}{!}{
    \begin{tabular}{cccccccccccc}
    \toprule
         Bit Precision&Method&Size(MB)&Avg&Bike&Lifestyle&Palace&Robot&Spaceship&Steamtrain&Toad&Wineholder  \\
         \midrule
         32-bit&KiloNeRF&$\leq$ 100&33.77&35.49&33.15&34.42&32.93&36.48&33.36&31.41&29.72\\
         8-bit$^\ast$&VM-192&17.77&36.11&38.69&34.15&37.09&37.99&37.66&37.45&34.66&31.16\\
         8-bit$^\ast$&VM-48&4.53&34.95&37.55&33.34&35.84&36.60&36.38&36.68&32.97&30.26\\
         8-bit$^\ast$&CP-384&0.72&33.92&36.29&32.29&35.73&35.63&34.58&35.82&31.24&29.75\\
         8-bit$^\ast$&VM-192 (300) + DWT&0.87&34.67&37.06&33.44&35.18&35.74&37.01&36.65&32.23&30.08\\
         \midrule
         8-bit$^\ast$&VM-192 (300) + Ours&8.98&36.24&38.78&34.21&37.22&38.02&38.61&37.79&34.39&30.97\\
         \bottomrule
         & 
    \end{tabular}
    }
    
    \label{tab:quant_nsvf}
\end{table*}

\subsection{LLFF Dataset}
\begin{figure*}
    \centering
    \includegraphics[width=\linewidth]{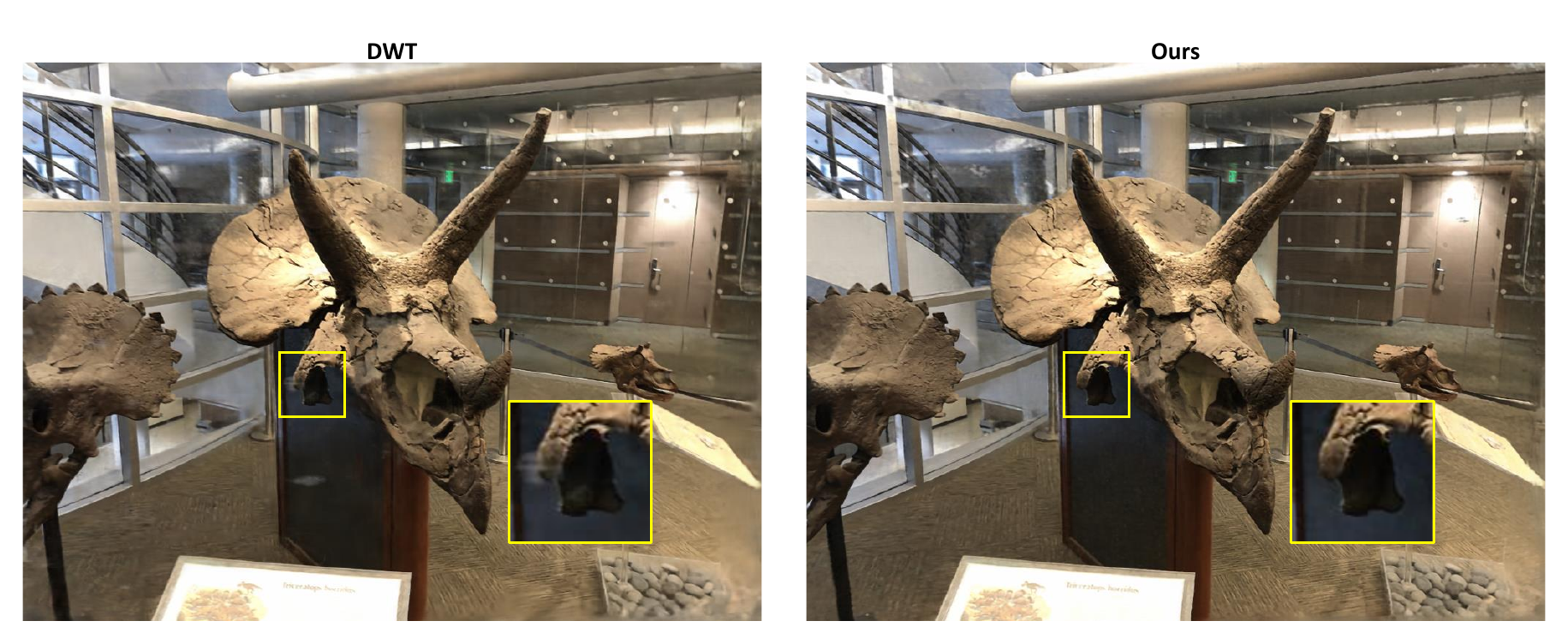}
    \caption{Visual comparison on LLFF synthetic dataset.}
    \label{fig:llff_1}
\end{figure*}

\begin{figure*}
    \centering
    \includegraphics[width=0.7\linewidth,height=\textheight]{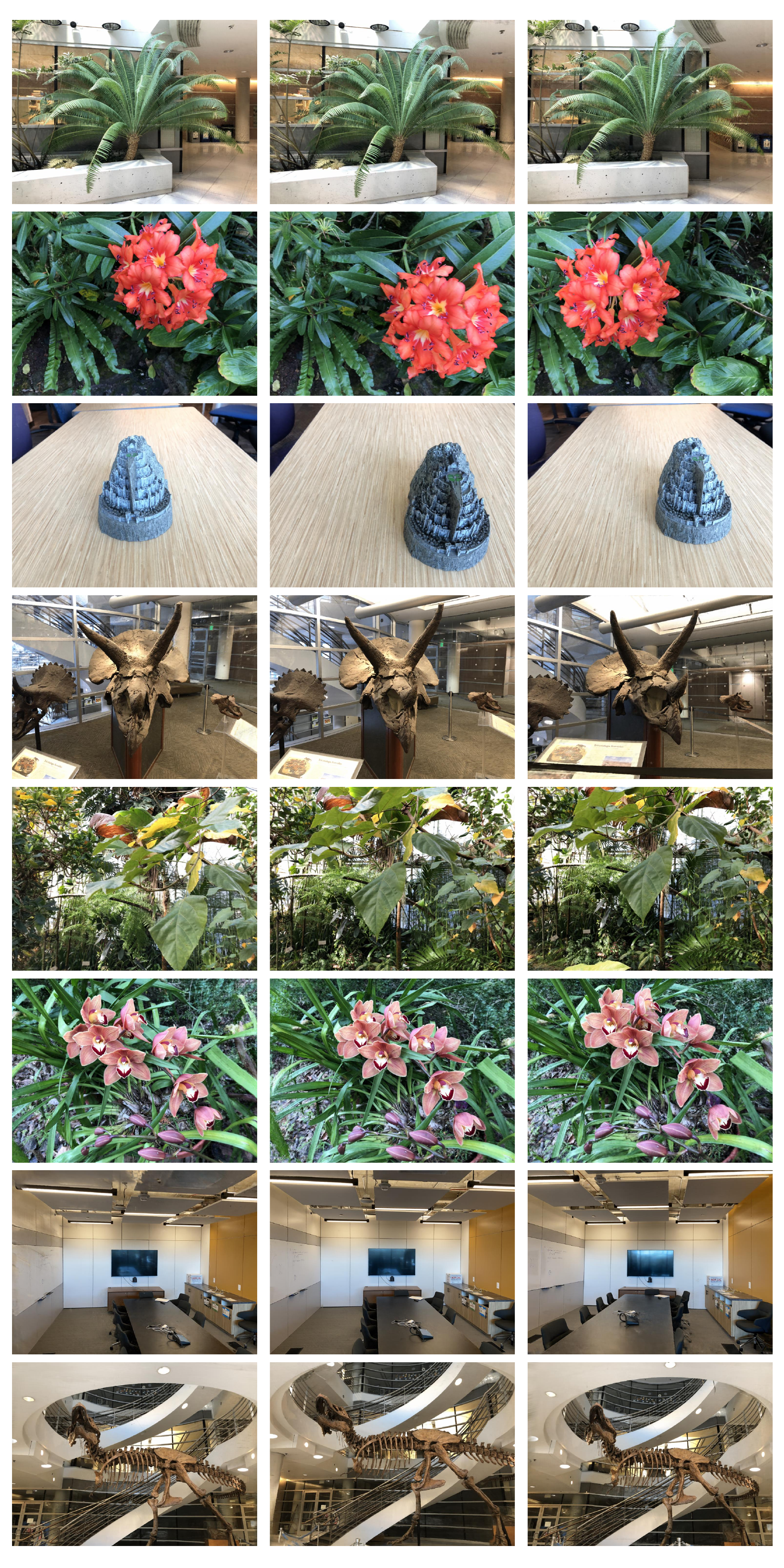}
    \caption{Visualizations on LLFF synthetic dataset.}
    \label{fig:llff_2}
\end{figure*}

The quantitative results for each case are presented in Table \ref{tab:quant_llff}, while additional visualizations comparing our representation with DWT \cite{rho2023masked} based representation method, are shown in Figure \ref{fig:llff_1}.  Furthermore, comprehensive visualizations of eight scenes in the NSVF dataset are shown in Figure \ref{fig:llff_2} and in the video.

\begin{table*}[ht]
    \centering
    \caption{Results of LLFF Dataset}
    \resizebox{\linewidth}{!}{
    \begin{tabular}{cccccccccccc}
    \toprule
         Bit Precision&Method&Size(MB)&Avg&Fern&Flower&Fortress&Horns&Leaves&Orchids&Room&T-Rex  \\
         \midrule
         8-bit&cNeRF&0.96&26.15&25.17&27.21&31.15&27.28&20.95&20.09&30.65&26.72\\
         8-bit$^\ast$&PREF&9.34&24.50&23.32&26.37&29.71&25.24&20.21&19.02&28.45&23.67\\
         8-bit$^\ast$&VM-96&44.72&26.66&25.22&28.55&31.23&28.10&21.28&19.87&32.17&26.89\\
         8-bit$^\ast$&VM-48&22.40&26.46&25.27&28.19&31.06&27.59&21.33&20.03&31.70&26.54\\
         8-bit$^\ast$&CP-384&0.64&25.51&24.30&26.88&30.17&26.46&20.38&19.95&30.61&25.35\\
         8-bit$^\ast$&VM-192 (300) + DWT&1.34&25.88&24.98&27.19&30.28&26.96&21.21&19.93&30.03&26.45\\
         \midrule
         8-bit$^\ast$&VM-192 (300) + Ours&13.67&26.48&25.02&28.23&31.07&27.81&21.24&19.68&31.82&26.97\\
         \bottomrule
         & 
    \end{tabular}
    }
    
    \label{tab:quant_llff}
\end{table*}

\section{Additional Ablation Studies}

\subsection{Sparsity Masks}
\begin{figure*}
    \centering
    \includegraphics[width=0.7\linewidth,height=\textheight]{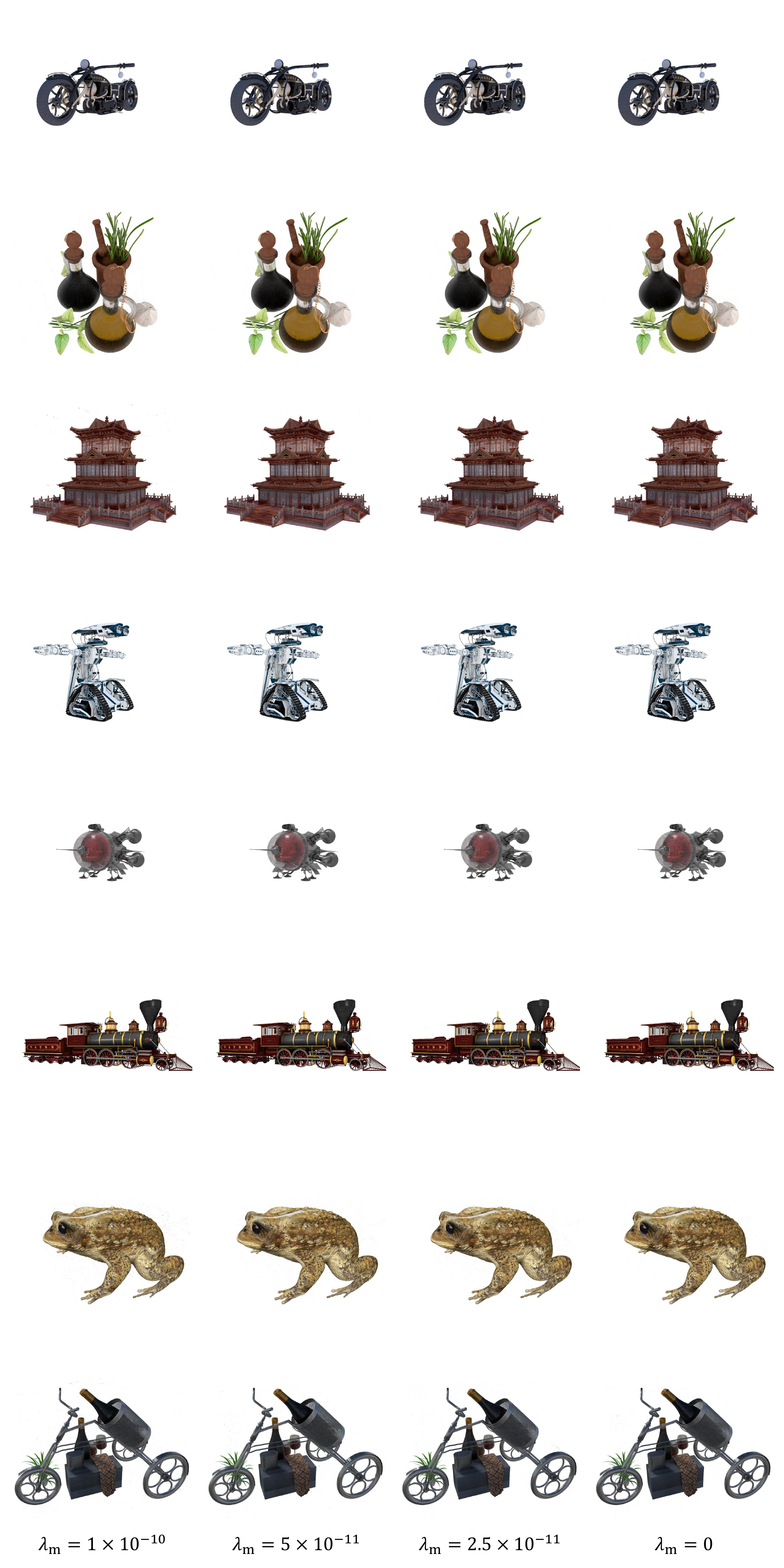}
    \caption{Qualitative results on NSVF dataset with different sparsity.}
    \label{fig:abl}
\end{figure*}

 We evaluate the performance of our direction-aware representation at various sparsity levels controlled by the mask loss weight $\lambda_m$. The quantitative and qualitative results on the NSVF dataset with different sparsity levels are presented in Table \ref{tab:abl_sparsity} and Figure \ref{fig:abl}.

 \begin{table*}[ht]
    \centering
    \caption{Quantitative results on NSVF dataset with different sparsity.}
    \resizebox{\linewidth}{!}{
    \begin{tabular}{cccccccccccc}
    \toprule
         Sparsity&$\lambda_m$&Size(MB)&Avg&Bike&Lifestyle&Palace&Robot&Spaceship&Steamtrain&Toad&Wineholder  \\
         \midrule
         99.2\%&$1.0\times10^{-10}$&1.16&35.36&38.01&33.69&35.70&37.23&37.83&37.26&32.58&30.56\\
         97.3\%&$5.0\times10^{-11}$&3.18&35.81&38.52&34.01&36.33&37.79&38.22&37.46&33.33&30.82\\
         94.2\%&$2.5\times10^{-11}$&8.98&36.24&38.78&34.21&37.22&38.02&38.61&37.79&34.39&30.97\\
         -&0&135&36.34&38.86&34.37&37.25&38.06&38.72&37.89&34.46&31.09\\
         \bottomrule
         & 
    \end{tabular}
    }
    \label{tab:abl_sparsity}
\end{table*}

\subsection{Wavelet Levels} 
We investigated the impact of scene reconstruction performance across different wavelet levels, and the results are presented in Table \ref{tab:wavelet_level}. Interestingly, we observed that increasing the wavelet level did not lead to significant performance improvements. Conversely, we noted a substantial increase in both training time and model size with the increment of wavelet level. As a result, throughout all experiments, we consistently set the wavelet level to 1.

\begin{table}[t]
    \centering
    \caption{
    \textbf{Wavelet Level Analysis of Direction-Aware Representation}, evaluated on NVSF data.}
    \resizebox{\linewidth}{!}{
    \begin{tabular}{cccc}
    \toprule
         Level&PSNR $\uparrow$&Model Size (MB)  $\downarrow$ &Training Time (min) $\downarrow$ \\
         \midrule
         1&36.34& \textbf{135}&\textbf{23}\\
         2&36.45&152&41\\
         3&\textbf{36.49}& 163&55\\
         \bottomrule
    \end{tabular}
    }
    \label{tab:wavelet_level}
\end{table}

\subsection{Training Time Analysis} 
To effectively demonstrate the efficiency of our proposed DaReNeRF, we conducted a comparative analysis against HexPlane under identical training durations of 2 hours (equivalent to HexPlane-100k) and 12 hours (equivalent to HexPlane-650k). The results, outlined in Table \ref{tab:traing_same_times}, reveal that across varying training periods, our proposed DaReNeRF consistently outperforms the baseline HexPlane.

\begin{table}[ht]
    \centering
    \caption{Time eval.\ on Plenoptic {\small(\textcolor{red}{\scriptsize\texttt{FlameSteak}}/\textcolor{blue}{\scriptsize\texttt{CutRoastBeef}})}.}
    \vspace{-.8em}
    \setlength{\tabcolsep}{1.5mm}
    \resizebox{\linewidth}{!}{
    \begin{tabular}{@{\hspace{.0em}}c@{\hspace{.2em}}|@{\hspace{.2em}}c@{\hspace{.2em}}c|c@{\hspace{.2em}}c|c@{\hspace{.2em}}c|c@{\hspace{.2em}}c|c@{\hspace{.2em}}c|c@{\hspace{.2em}}c@{\hspace{.0em}}}
    \toprule
         \multirow{2}*{Model}&\multicolumn{6}{c|}{Eval.\ after training for \textbf{2hrs}}&\multicolumn{6}{c}{Eval.\ after training for \textbf{12hrs}} \\
         & \multicolumn{2}{c}{PSNR $\uparrow$} & \multicolumn{2}{c}{D-SSIM $\downarrow$} & \multicolumn{2}{c|}{LPIPS $\downarrow$} &
         \multicolumn{2}{c}{PSNR $\uparrow$} & \multicolumn{2}{c}{D-SSIM $\downarrow$} & \multicolumn{2}{c}{LPIPS $\downarrow$}\\
        \midrule
        HexPlane & \textcolor{red}{31.92} & / \textcolor{blue}{32.71} & \textcolor{red}{.012} & / \textcolor{blue}{.015} & \textcolor{red}{.081} & / \textcolor{blue}{.094}
        & \textcolor{red}{32.08} & / \textcolor{blue}{32.55} & \textcolor{red}{.011} & / \textcolor{blue}{.013} & \textcolor{red}{.066} & / \textcolor{blue}{.080}\\
        DaReNeRF & \textcolor{red}{\textbf{33.01}} & / \textcolor{blue}{\textbf{32.98}} & \textcolor{red}{\textbf{.010}} & / \textcolor{blue}{\textbf{.013}} & \textcolor{red}{\textbf{.079}} & / \textcolor{blue}{\textbf{.092}}
        & \textcolor{red}{\textbf{33.62}} & / \textcolor{blue}{\textbf{33.43}} & \textcolor{red}{\textbf{.009}} & / \textcolor{blue}{\textbf{.010}} & \textcolor{red}{\textbf{.063}} & / \textcolor{blue}{\textbf{.076}}\\

    \bottomrule
    \end{tabular}
    }
    \label{tab:traing_same_times}
\end{table}

\subsection{Training Data Sparsity Analysis} 
In order to delve deeper into the few-shot capabilities of our proposed direction-aware representation, we conducted experiments with varying levels of training data sparsity. This was achieved by randomly dropping training frames while ensuring sufficient data remained to effectively learn motion on the D-NeRF dataset. The corresponding results are presented in Table \ref{tab:sparsity_of_training_set}. Remarkably, our proposed DaReNeRF consistently outperforms the baseline across different levels of training data sparsity.
\begin{table}[h]
    \centering
    \setlength{\tabcolsep}{2.0mm}
    \caption{Evaluation on D-NeRF with various training set sparsity.}
    \resizebox{\linewidth}{!}{
    \begin{tabular}{c|ccc|ccc}
    \toprule
         \multirow{2}*{Model}&\multicolumn{3}{c|}{\textbf{75\%} training set (average)} &\multicolumn{3}{c}{\textbf{50\%} training set (average)} \\
         ~&PSNR $\uparrow$ &SSIM $\uparrow$ &LPIPS $\downarrow$ &PSNR $\uparrow$ &SSIM $\uparrow$ &LPIPS $\downarrow$\\
        \midrule
        HexPlane&29.85&0.95&0.05&28.03&0.94&0.06\\
        DaReNeRF&\textbf{30.95}&\textbf{0.96}&\textbf{0.04}&\textbf{29.28}&\textbf{0.96}&\textbf{0.05}\\

    \bottomrule
    \end{tabular}
    }
    \label{tab:sparsity_of_training_set}
\end{table}

\section{Training Details}
\subsection{Plenoptic Video Dataset \cite{li2022neural}}
Plenoptic Video Dataset is a multi-view real-world video dataset, where each video is 10-second long. For training, we set $R_1=48$, $R_2=48$ and $R_3=48$ for appearance, where $R_1$, $R_2$ and $R_3$ are basis numbers for direction-aware representation of $XY-ZT$, $XZ-YT$ and $YZ-XT$ planes. For opacity, we set $R_1=24$, $R_2=24$ and $R_3=24$. The scene is modeled using normalized device coordinate (NDC) \cite{mildenhall2021nerf} with min boundaries $[-2.5,-2.0,-1.0]$ and max boundaries $[2.5,2.0,1.0]$. 

During the training, DaReNeRF starts with a space grid size of $64^3$ and double its resolution at 20k, 40k and 70k to $512^3$. The emptiness voxel is calculated at 30k, 50k and 80k. The learning rate for representation planes is 0.02 and the learning rate for $V^{RF}$ and neural network is 0.001. All learning rates are exponentially decayed. We use Adam \cite{kingma2014adam} optimization with $\beta_1=0.9$ and $\beta_2=0.99$. We apply the total variational loss on all representation planes with loss weight $\lambda=1e-5$ for spatial planes and $\lambda=2e-5$ for spatial-temporal planes. For DaReNeRF-S we set weight of mask loss as $1e-11$. 

We follow the hierarchical training pipeline suggested in \cite{li2022neural}. Both DaReNeRF and DaReNeRF-S use 100k iterations, with 10k stage one training, 50k stage two training and 40k stage three training. Stage one is a global-median-based weighted sampling with $\gamma=0.02$; stage two is also a global-median based weighted sampling with $\gamma=0.02$; stage three is a temporal-difference-based weighted sampling with $\gamma=0.2$.

In evaluation, D-SSIM is computed as $\dfrac{1-MS-SSIM}{2}$ and LPIPS \cite{zhang2018unreasonable} is calculated using AlexNet \cite{krizhevsky2012imagenet}. 

\subsection{D-NeRF Dataset \cite{pumarola2021d}}
We set $R_1=48$, $R_2=48$ and $R_3=48$ for appearance and $R_1=24$, $R_2=24$ and $R_3=24$ for opacity. The bounding box has max boundaries $[1.5,1.5,1.5]$ and min boundaries $[-1.5,-1.5,-1.5]$. During the training, both DaReNeRF and DaReNeRF-S starts with space grid of $32^3$ and upsampling its resolution at 3k, 6k and 9k to $200^3$. The emptiness voxel is calculated at 4k, 8k and 10k iterations. Total training iteration is 25k. The learning rate for representation planes are 0.02 and learning rate for $V^{RF}$ and neural network is 0.001. All learning rates are exponentially decayed. We use Adam \cite{kingma2014adam} optimization with $\beta_1=0.9$ and $\beta_2=0.99$. In evaluation, LPIPS \cite{zhang2018unreasonable} is calculated using VGG-Net  \cite{simonyan2014very} following previous works. 

For \textbf{both} the Plenoptic Video dataset and the D-NeRF dataset, we set the learning rate of the masks in DaReNeRF-S same as their representation planes and we employ a compact MLP for regressing output colors. The MLP consists of 3 layers, with a hidden dimension of 128.

\subsection{Static Scene}
For three static scene dataset NeRF synthetic dataset, NSVF synthetic dataset and LLFF dataset, we followed the experimental settings of TensoRF \cite{chen2022tensorf}. We trained our model for 30000 iterations, each of which is a minibatch of 4096 rays. We used Adam \cite{kingma2014adam} optimization with $\beta_1=0.9$ and $\beta_2=0.99$ and an exponential learning rate decay scheduler. The initial learning rates of representation-related parameters and neural network (MLP) were set to 0.02 and 0.001. For the \textbf{NeRF synthetic} and \textbf{NSVF synthetic} datasets, we adopt TensoRF-192 as the baseline and update the alpha masks at the 2k, 4k, 6k, 11k, 16k, and 26k iterations. The initial grid size is set to $128^3$, and we perform upsampling at 2k, 3k, 4k, 5.5k, and 7k iterations, reaching a final resolution of $300^3$. For the \textbf{LLFF} dataset, we adopt TensoRF-96 as the baseline and update the alpha masks at the 2.5k, 4k, 6k, 11k, 16k, and 21k iterations. The initial grid size is set to $128^3$, and we perform upsampling at 2k, 3k, 4k and 5.5k iterations, reaching a final resolution of $640^3$. The learning rates of masks are set same as learning rates of representation-related parameters. We employ a compact MLP for regressing output colors. The MLP consists of 3 layers, with a hidden dimension of 128.